\documentclass[lettersize,journal]{IEEEtran}
\usepackage{amsmath,amsfonts}
\usepackage{algorithmic}
\usepackage{algorithm}
\usepackage{array}
\usepackage[caption=false,font=normalsize,labelfont=sf,textfont=sf]{subfig}
\usepackage{textcomp}
\usepackage{stfloats}
\usepackage{url}
\usepackage{verbatim}
\usepackage{graphicx}
\usepackage{cite}
\newtheorem{assump}{Assumption}

\newtheorem{thm}{Theorem}

\newtheorem{prop}{Property}
\newtheorem{proof}{Proof}
\begin{document}

\title{Convex Latent-Optimized Adversarial Regularizers for Imaging Inverse Problems}

\author{Huayu Wang, Chen Luo, Taofeng Xie, Qiyu Jin, Guoqing Chen, Zhuo-Xu Cui, Dong Liang~\IEEEmembership{Senior Member,~IEEE}
\thanks{This work was supported in part by the National Key R$\&$D Program of China (2021YFF0501503, 2020YFA0712202 and 2022YFA1004202); National Natural Science Foundation of China (U21A6005, 62125111, 12026603, 62206273, 61771463, 81830056, U1805261, 81971611, 61871373, 81729003, 81901736); Key Laboratory for Magnetic Resonance and Multimodality Imaging of Guangdong Province (2020B1212060051).}
\thanks{Manuscript received April 19, 2021; revised August 16, 2021.}
\thanks{Corresponding author: zx.cui@siat.ac.cn and dong.liang@siat.ac.cn}
\thanks{H. Wang and C. Luo contributed equally to this work}
\thanks{H. Wang, C. Luo, T. Xie, Q. Jin and G. chen are with School of Mathematical Sciences, Inner Mongolia University, Hohhot, China.}
\thanks{Z.-X. Cui and D. Liang are with Research Center for Medical AI, Shenzhen Institutes of Advanced Technology, Chinese Academy of Sciences, Shenzhen, China.}}
\markboth{Journal of \LaTeX\ Class Files,~Vol.~14, No.~8, August~2021}%
{Shell \MakeLowercase{\textit{et al.}}: A Sample Article Using IEEEtran.cls for IEEE Journals}


\maketitle

\begin{abstract}
Recently, data-driven techniques have demonstrated remarkable effectiveness in addressing challenges related to MR imaging inverse problems. However, these methods still exhibit certain limitations in terms of interpretability and robustness. In response, we introduce Convex Latent-Optimized Adversarial Regularizers (CLEAR), a novel and interpretable data-driven paradigm. CLEAR represents a fusion of deep learning (DL) and variational regularization. Specifically, we employ a latent optimization technique to adversarially train an input convex neural network, and its set of minima can fully represent the real data manifold.
We utilize it as a convex regularizer to formulate a CLEAR-informed variational regularization model that guides the solution of the imaging inverse problem on the real data manifold. Leveraging its inherent convexity, we have established the convergence of the projected subgradient descent algorithm for the CLEAR-informed regularization model. This convergence guarantees the attainment of a unique solution to the imaging inverse problem, subject to certain assumptions.
Furthermore, we have demonstrated the robustness of our CLEAR-informed model, explicitly showcasing its capacity to achieve stable reconstruction even in the presence of measurement interference. Finally, we illustrate the superiority of our approach using MRI reconstruction as an example. Our method consistently outperforms conventional data-driven techniques and traditional regularization approaches, excelling in both reconstruction quality and robustness.
\end{abstract}


\section{Introduction}
Magnetic Resonance Imaging (MRI)\cite{lauterbur1973image} is a widely employed imaging method in clinical applications. It excels at generating high-resolution images and offers the advantages of being non-invasive and radiation-free, thereby minimizing its impact on patients. As a result, MRI finds widespread usage in various clinical treatment scenarios. However, MRI does have certain limitations. The signal acquisition process is time-consuming, and both the subtle movements of the patient's body and the physiological motion of body tissues during acquisition can adversely affect the obtained results and reduce imaging quality. Consequently, the urgent challenge lies in accelerating the imaging speed of MRI.

Previous studies have devoted considerable efforts to accelerating MRI from multiple perspectives, including enhancements in imaging hardware \cite{sodickson1997simultaneous}, the design of improved imaging sequences \cite{kaiser1989mr}. Moreover, undersampling emerges as the most straightforward approach to achieving acceleration. However, the task of reconstructing high-quality images from undersampled data resides in addressing ill-posed problems. Consequently, several hand-crafted regularization methods for accelerated MRI have been proposed \cite{lustig2008compressed,ma2008efficient,song2020score}.

In recent years, data-driven methodologies based on deep learning (DL) have exhibited remarkable efficacy across diverse image processing domains \cite{ronneberger2015u,wang2022dimension,an2015variational}, including MRI \cite{cui2023k,10177777}. Nonetheless, these methods involve direct learning of the mapping between undersampled data and accurate images \cite{goodfellow2014generative,zhu2017unpaired}, yet they may lack a certain level of interpretability. Furthermore, the presence of nonconvex and nonlinear complexities within deep network architectures poses a challenge to maintaining imaging stability. Frequently, even slight perturbations in the measurement engender noteworthy deterioration in the reconstructed outcomes \cite{antun2020instabilities}.

To address the aforementioned limitations, a learnable regularization approach that integrates DL with regularization models has been introduced \cite{aggarwal2018modl,zhang2018ista,ho2020denoising}. In particular, Lunz et al.\cite{lunz2018adversarial} proposed Adversarial Regularizers (AR) trained by neural networks. Their work aimed to learn regularizers that represent the distance between the initial input and the real image manifold through adversarial learning. Then, based on \cite{lunz2018adversarial}, Mukherjee et al. utilized the input convex neural networks\cite{amos2017input} to obtain the regularizers resulting in global convergence for its iterative algorithm. However, the regularizers derived from the aforementioned approach can only measure the distance between the initial input and the real image manifold. As the iterations progress, the latent iterations may fail to align with the relationship between the initial input and the real data distribution. Consequently, this approach is unable to gauge the distance between latent iterations and the true data manifold, ultimately leading to errors in the iterative reconstruction.

\subsection{Contributions}
To address this constraint, the concept of latent-optimization is incorporated into the training process of the regularizer. This innovation enables the regularizer that guides the latent iterate towards the true data manifold. Such an enhancement not only elevates the model's interpretability but also improves the accuracy of reconstruction outcomes. In this paper, we have introduced a learnable convex regularization technique, denoted as Convex Latent-optimizEd Adversarial Regularizers (CLEAR), characterized by its interpretability and stability. The principal contributions of this study are as follows:

\begin{itemize}
    \item For imaging inverse problems, we proposed a learnable convex adversarial regularizer based on latent-optimization, referred to as CLEAR, and its set of minima can fully represent the real
    data manifold. Furthermore, we utilize it as a convex regularizer to formulate a CLEAR-informed variational regularization model that guides the solution of the imaging inverse problem on the real data manifold.
    
    \item With the inherent convexity being leveraged, the convergence of the projected subgradient descent algorithm for the CLEAR-informed regularization model was demonstrated. This convergence leads to a unique solution of the imaging inverse problem, under some assumptions. 
    
    \item The robustness of our CLEAR-informed regularization model is demonstrated, with stable reconstruction being achieved even in the presence of certain measurement interference.

    \item Experiments conducted on two MR datasets conclusively demonstrate the remarkable superiority of the proposed CLEAR over traditional TV methods, the AR-learned regularization approach, and WGAN, both in terms of image reconstruction quality and resilience against noisy interference. Importantly, it is worth highlighting that CLEAR derives its advantage from its convex nature, thus mitigating the risk of getting trapped in local minima. CLEAR consistently outperforms its non-convex ablation counterpart in terms of reconstruction quality and robustness.

\end{itemize}

The remainder of the paper is organized as follows. Section \ref{sec2} introduces the background knowledge and related works.  Section \ref{sec3} presents the proposed CLEAR and its theoretical properties. Section \ref{sec4} describes the training process of CLEAR. The implementation details are presented in Section \ref{sec5}. Experiments performed on several datasets are presented in Section \ref{sec6}. A discussion is presented in Section \ref{sec7}. Section \ref{sect8} provides some concluding remarks. All proofs are presented in the appendix.

\section{Background $\&$ Related Works}\label{sec2}
\subsection{Variational Regularization}\label{sec2.1}
In many imaging problems, the forward models can be expressed as follows, without considering noise perturbations:
\begin{equation}\label{forward}Ax = b
\end{equation}
where $A$ is the imaging system, $x$ is the sought image and $b$ is the measurement.
Its inverse problem is often ill-posed, making it challenging to directly obtain $x$ through inverse operations. To address this issue, a suitable regularization term is introduced, transforming the inverse problem into the following optimization form:$$\arg\min\limits_{x}{\left \| Ax-b \right \|^2 +\lambda R(x)}$$
where the first term represents the data fidelity term, ensuring the solution aligns with the measured data. The second term is the regularizer, which represents the image priors. The solution to the optimization problem is then obtained using an iterative method such as gradient descent, proximal gradient descent, etc.

A successful regularization model should possess the following attributes: it should be underpinned by solid mathematical and physical principles to ensure the interpretability of the solution; the corresponding iterative algorithm should converge towards the solution of the inverse problem; and the solution should exhibit robustness to noise perturbations in the measurements.

\subsection{Adversarial Regularizers (AR)}
To find a suitable regularizer, Lunz et al. introduced the learnable adversarial regularizers method \cite{lunz2018adversarial}. They assumed that the true image or data lies on a manifold and follows the distribution $\mathbb{P}_r$, while the input data lies on another data manifold and follows the distribution $\mathbb{P}_{n}$.
Ideally, it is desirable to identify a distance metric that serves as a regularizer to measure the distance between two manifolds. This approach aims to achieve the projection of the initial input onto the real data manifold by minimizing this distance metric. To obtain the distance function, Lunz et al. introduced an adversarial learning model:
\begin{equation}\label{adversarial}
    \max\limits_{f\in \text{1-Lip}  }{\mathbb{E}_{x\sim \mathbb{P}_{n} }\left [ f\left (x \right )  \right ]  - \mathbb{E}_{x\sim \mathbb{P}_{r} }\left [ f\left ( x \right )  \right ] }
\end{equation}
where $\text{1-Lip}$ represents the set of bounded 1-Lipschitz continuous functions. Practically, they represented function $f$ by the neural networks $\Psi _{\Theta }\left ( x \right ) $ and trained adversarially with the Gradient Penalty (GP) loss \cite{NEURIPS2018_91d0dbfd}.
If $\Psi _{\Theta^*} $ attains the maximum of (\ref{adversarial}), Lunz et al. have demonstrated that $\Psi _{\Theta^*} $ serves as a distance metric measuring the distance from the input data manifold to the real data manifold.

However, the regularizers derived from the aforementioned approach can only measure the distance between the initial input and the real image manifold. As the iterations progress, the latent iterations may fail to align with the relationship between the initial input and the real data manifold. Therefore this method poses drawbacks in terms of interpretability. Additionally, Lunz et al. did not provide a complete proof of convergence for their proposed method. 
\subsection{Learned Convex Regularizers}\label{sec2.3}
Recently, \cite{amos2017input} showed that the neural network can be modified to maintain its convexity with respect to the input. Mukherjee et al. utilized a convex network architecture to realize the AR \cite{mukherjee2020learned}.
Due to convexity, the convergence of the algorithm for solving the regularization problem can be ensured. However, constrained by the adversarial learning model (\ref{adversarial}), after a one-step iteration, a deviation in learned distance measurement introduced by the AR model still persists.

Specifically, the convexity of the neural network is preserved through appropriate weight clipping, based on the following two facts\cite{boyd2004convex}:
\begin{itemize}
	\item Non-negative combination of finitely many convex functions is another convex function, and
 \item The composition $\psi _{1} \circ\psi _{2}$ of two function $\psi _{1}$ and $\psi _{2}$ is convex when $\psi _{2}$ is convex and $\psi _{1}$ is convex and monotonically non-decreasing
 \end{itemize}

\section{Methodology and Theory}\label{sec3}
\subsection{Interpretable Learnable Regularizers}
In this section, to address the issue of deviation from the learned distance metric after a one-step iteration of AR, we propose a Convex Latent-optimizEd Adversarial Regularizers (CLEAR) model. Specifically it takes the following optimization form:
\begin{equation}\label{eq1}\max\limits_{f\in \Gamma } {\min_{x \in X}{f\left ( x \right ) } - \mathbb{E} _{ x^{+}\sim \mathbb{P} _{r} }\left[ f\left (  x^{+}\right ) \right ]}\end{equation}
where $\Gamma$ represents the set of bouned 1-Lipschitz continuous convex functions, $\mathbb{P} _{r}$ represents the distribution of real data $\left \{ x^{+} \right \} $.

In response to AR model (\ref{adversarial}), we have implemented two key enhancements:
\begin{itemize}
\item \textbf{Latent Optimization}: Intuitively, instead of learning the distance from input data to the true data manifold, we now learn the distance between the optimal solution of $f$ and the true data manifold. Consequently, the learned function $f$ becomes independent of the input data; as long as its optimal value can be determined, it can effectively measure the distance to the real data manifold. When this well-learned $f$ is incorporated into the regularization model and minimized, its solution is compelled to lie on the real data manifold.

\item \textbf{Convexity Constraint}: We impose a constraint on $f$, requiring it to be a convex function. This constraint ensures that the optimal value of $f$ can be readily ascertained, contributing to the effectiveness of the method.
 \end{itemize}

Next, we explore the theoretical potential of the function $f$ obtained from the CLEAR model (\ref{eq1}) to serve as a ``distance". To begin, we establish precise definitions for the real data manifold and distance.
\begin{assump}\label{assup1}
\textit{Assume that the real data is sampled from a distribution $\mathbb{P}_r$ which is supported on the convex compact set $\mathcal{M}$ of $X$, i.e., $\mathbb{P}_r(\mathcal{M}^c)=0$.}
\end{assump}

Assumption \ref{assup1} is a common assumption and has been used in \cite{}. Under Assumption \ref{assup1}, we denote $d_{\mathcal{M}}(x)$ as the distance from sample $x$ to $\mathcal{M}$. $d_{\mathcal{M}}(x)$ is defined as follows,
$$d_{\mathcal{M} }\left ( x \right ) =\min_{x'\in \mathcal{M} } {d\left ( x,x' \right )}, \quad d\left ( x,x' \right )=\left \| x-x' \right \|$$
If with no special statement, in this paper, $\left \| \cdot  \right \| $ represents the $L_2$-norm.
By the definition of $d_{\mathcal{M} }\left ( x \right ) $, we can get a property of $d_{\mathcal{M} }\left ( x \right ) $ based on Assumption\ref{assup1}.
\begin{prop}\label{prop1}
\textit{$d_{\mathcal{M} }\left ( x \right ) $ is convex and 1-Lipschitz continuous.}
\end{prop}
Related proof can be found in \cite{cui2022deep} Theorem3.3.

Based on the above assumption, we have the following result for the CLEAR model.
\begin{thm}\label{thm2}
Suppose Assumption \ref{assup1} holds. If $f^*$ is the maximum of CLEAR model (\ref{eq1}) then, $\Omega = \mathcal{M}$, a.e., where $\Omega := \{ x \in X|f^*\left( x \right) =\min_{x\in X}f^*\left ( x \right )    \} $ and $X$ denotes the finite-dimensional space in which the samples are located.
\end{thm}

The detailed proof is presented in the Appendix. The above theorem states that two sets $\Omega$ and $\mathcal{M}$ are equivalent almost everywhere (a.e.). 
While $f^*$ is not entirely equivalent to $d_{\mathcal{M} }$, it can serve the role of $d_{\mathcal{M}}$. 
For all samples $x^{*}$ in $\Omega$, the distance from $x^{*}$ to $\mathcal{M}$ is zero $d_{\mathcal{M}}\left(x^{*}\right) = 0$. Similarly, for all samples $x^{+}$ in $\mathcal{M}$, the function $f^*$ maps $x^{+}$ to the minimum value of $f^*$ over the entire space $X$, i.e., $f^*\left(x^{+}\right) = \min_{x\in X}f^*\left(x\right)$. Throughout the rest of this paper, we will consider the almost everywhere scenario, where $\Omega$ and $\mathcal{M}$ can be regarded as the same set due to their equivalence a.e.. With $f^*\left(x\right)$ defined as such, finding samples that minimize it is equivalent to finding samples on the true data manifold $\mathcal{M}$, and the proposed regularizer is defined as such $f^*\left(x\right)$.

For inverse problem (\ref{forward}), CLEAR-informed variational regularization model takes the form:  
\begin{equation}\label{eq4}\min_{x\in X}{\left \| Ax-b \right \|^2 +\lambda f^*\left ( x \right )}.\end{equation}

Then, let us state an assumption on the learned regularizer $f^*$ that aligns with inverse problem (\ref{forward}).
\begin{assump}\label{assup2}
\textit{The real data manifold $\mathcal{M}$ intersects the solution set of inverse problem of $Ax=b$ with a unique real solution $x^{+}$, i.e., $\mathcal{M}\cap \mathcal{A} = \left \{ x^{+} \right \} $, $\mathcal{A} = \left \{x\in X|Ax=b  \right \}$}
\end{assump}

Under the Assumptions \ref{assup1} and \ref{assup2}, we have the following property.
\begin{thm}\label{prop2}
Suppose Assumptions \ref{assup1} and \ref{assup2} hold, and the maximum of the CLEAR model (\ref{eq1}), denoted as $f^*$, can be determined. Then, the real solution $x^{+}$ represents the unique minimum of the CLEAR-informed variational regularization model (\ref{eq4}).
\end{thm}

The intersection point $x^{+}$ lies on $\mathcal{M}$, which, according to Theorem \ref{thm2}, means $x^{+} \in \Omega$. Therefore, $f\left ( x^{+} \right )$ is minimized. Additionally, it satisfies the equation $Ax=b$, ensuring that the first term on the right-hand side of (\ref{eq4}) achieves the minimum value 0. Therefore, $x^{+}$ is the unique minimum of (\ref{eq4}). In MR imaging inverse problem, the full sampling image is the unique $x^{+}$ corresponding to $Ax=b$, which means that we can reconstruct it well via regularization problem (\ref{eq4}).

\subsection{Convergence}
After training the CLAER model, we utilized the CLEAR-informed regularization method to solve the inverse problem, which involves solving the variational regularization problem (\ref{eq4}). In this paper, we employed the projected (sub)gradient descent method (PGD) \cite{boyd2003subgradient} to solve it, which reads: 
\begin{algorithm}[H]
\caption{Projected Subgradient Descent (PGD)}\label{alg:pgd}
\begin{algorithmic}
\STATE 
\STATE $\textbf{input } x_{0}, f^*, \left \{ t_{i} \right \}, k, \text{projection on~} \mathcal{A}: \mathcal{P}_{\mathcal{A}}   $ 
\STATE $\textbf{for i = 1:k, loop:}$
\STATE \hspace{0.5cm}$x_{i-0.5} = x_{i-1}-t_{i-1} \cdot  \partial f^*\left ( x_{i-1} \right )$
\STATE \hspace{0.5cm}$x_{i} = \mathcal{P}_{\mathcal{A}}\left (x_{i-0.5}   \right ) $
\STATE \hspace{0.5cm}$\textbf{end}$
\STATE $\textbf{output } x_{k}$
\end{algorithmic}
\label{alg1}
\end{algorithm}

Supposed the above assumptions hold, we will show that the PGD algorithm converges to the real solution $x^{+}$ of inverse problem (\ref{forward}).

\begin{thm}\label{thm4}
Suppose Assumptions \ref{assup1} and \ref{assup2} hold, and the maximum of the CLEAR model (\ref{eq1}), denoted as $f^*$, can be determined. Then, the iterates generated by Algorithm \ref{alg:pgd} converge to unique real solution $x^{+}$ of inverse problem (\ref{forward}).
\end{thm}

A detailed proof is presented in the Appendix.

\subsection{Stabilty}

During the data acquisition, measurements are inevitably contaminated by noise, making it essential for the reconstruction method to be robust to noise. When the data is affected by small amounts of noise, the optimal solution should not undergo drastic changes. Mathematically, this implies that the solution should be continuous with respect to the measurements. Next, we will demonstrate that the proposed method possesses this property.
\begin{thm}\label{thm5}
Suppose Assumptions \ref{assup1} and \ref{assup2} hold, and the maximum of the CLEAR model (\ref{eq1}), denoted as $f^*$, can be determined. 
When measurements are perturbed by noise with level $\delta$, which is denoted as $b_{\delta}$, and the corresponding PGD iteration is denoted as $x^{k}_{\delta}$, then, $x_{\delta}^k \to x^{+}$, as $\delta \to 0, k\to\infty$.
\end{thm}

Further details of the proof can be found in the Appendix. Theorem \ref{thm5} demonstrates that when the measurement is contaminated by small noise, the reconstruction result of the proposed method undergoes only small changes, indicating its robustness to measurement noise. 

\section{Traing Procedure}\label{sec4}

In this section, we will elaborate on the training process of our proposed method.

To begin, we focus on training the learnable convex regularizers that were discussed in the preceding section. Our objective is to find a suitable function $f^{*}\left ( x \right )$ as defined in Section \ref{sec3}. We achieved it by employing the convex neural network $\Phi \left ( x;\theta \right )$ as the representation of $f^{*}\left ( x \right )$.

Based on the principles mentioned in Section \ref{sec2.3}, we designed the network architecture of $\Phi \left ( x;\theta \right )$ to maintain its convexity. Starting from a sample $x_{0}$, we used convex optimization methods to find the corresponding sample $x^{*}$ that belongs to the set $\Omega_{\theta}$, where $\Omega_{\theta} = \left \{ x|\Phi \left ( x;\theta  \right ) =\min_{x\in X}\Phi \left ( x;\theta  \right )   \right \} $. In this study, we applied significant levels of noise interference to real images $x^{+}$ to generate the input images $x_{0}$. During the training process, we employed stochastic gradient descent method (SGD) to solve $\min\limits_{x\in X}\Phi \left ( x;\theta \right )$, which can be expressed as:
\begin{align}\label{eq8}
x_{0} =& x^{+}+\delta_{0},\\
x_{k} =& x_{k-1}-\eta \cdot \bigtriangledown_{x}\Phi \left ( x;\theta  \right )+\delta_{k},\quad k=1,2,\ldots,t,\\
x^{*} =& x_{t}
\end{align}
where $\eta$ denotes the descent step, $\bigtriangledown_{x}$ represents the (sub)gradient, $\delta_{0}$ represents the significant levels of noise, and $\delta_{k}$ represents small random noise. The parameter $t$ represents the maximum number of iterations.

Due to the convexity of the network, this generation process will not get stuck in a local minimum, allowing us to find $\min\limits_{x\in X}\Phi \left ( x;\theta \right )$ more accurately.

\begin{figure}[!t]
\centering
\includegraphics[width=3.5in]{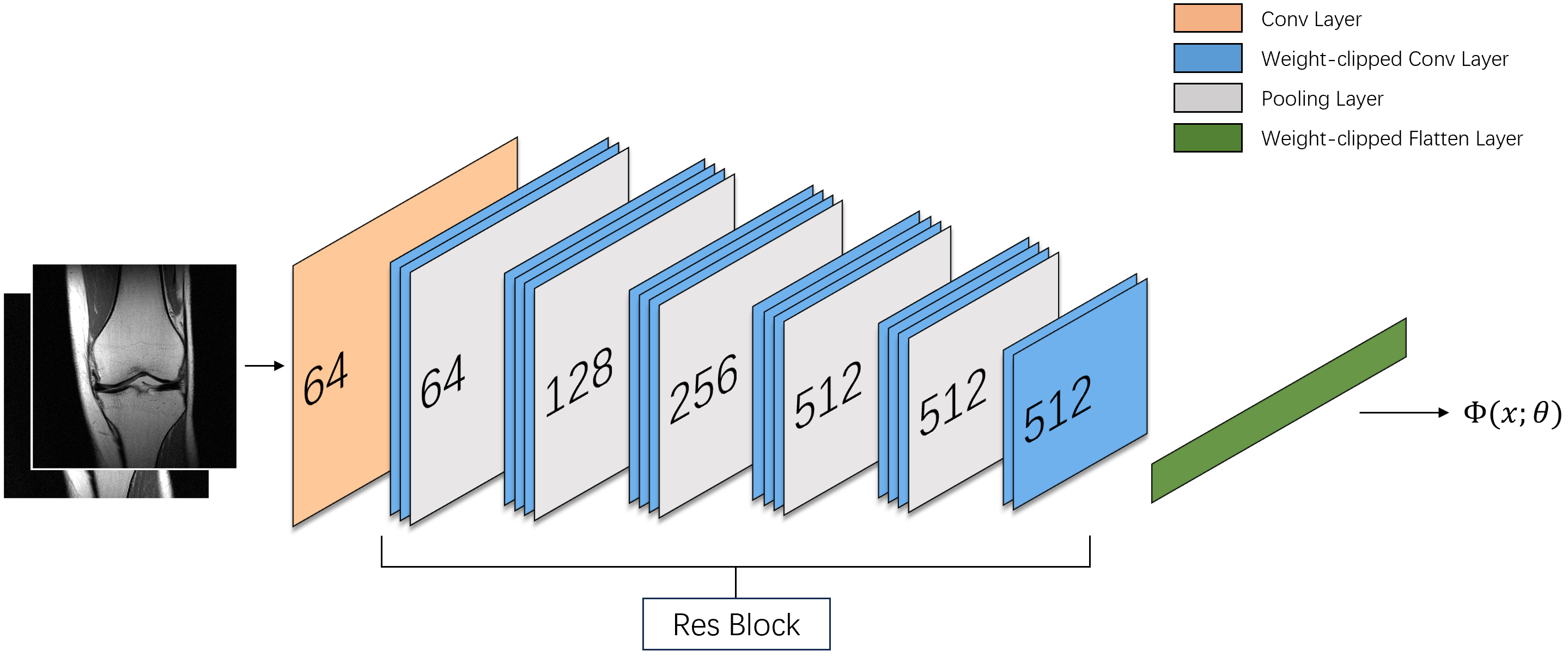}
\caption{Convex network architecture used in this study. The orange convolutional layer is left untreated, while the blue convolutional layers and green fully connected layer undergo weight clipping, retaining only non-negative weights.}
\label{net_stru}
\end{figure}

We attempted to train the eligible $\Phi \left ( x;\theta \right )$ using the loss function of the neural network. The expectation in (\ref{eq1}) is replaced with a simple average of the training samples to obtain the loss function. It is worth noting that $f\left ( x \right )$ is 1-Lipschitz continuous. In similar works, it is a common practice to include a gradient penalty term in the loss function to ensure that the network remains 1-Lipschitz continuous. The final loss function used to train the network is as follows:
\begin{equation}
\begin{aligned}\label{eq11}
\mathfrak{L}\left ( x^{+};\theta \right )=& \frac{1}{n}\sum_{i=1}^{n}{\Phi\left ( x^{+}_{i};\theta  \right )}-  \frac{1}{n}\sum_{i=1}^{n}{\Phi\left ( x^{*}_{i};\theta  \right )}\\& + \lambda\cdot  \mathbb{E}\left [ \left (\left \|    \bigtriangledown _{x} \Phi\left ( x;\theta  \right )\right \| -1\right   )^{2}  \right ] 
\end{aligned}
\end{equation}
After comparing several network architectures, we selected Res-Net\cite{he2016deep} as the network architecture to represent the regularizers, which has demonstrated excellent performance in various image processing tasks\cite{zhang2017beyond}\cite{zhang2019attention}. The network structure is shown in Fig.\ref{net_stru}.

\begin{figure}[!t]
\centering
\includegraphics[width=1.3in]{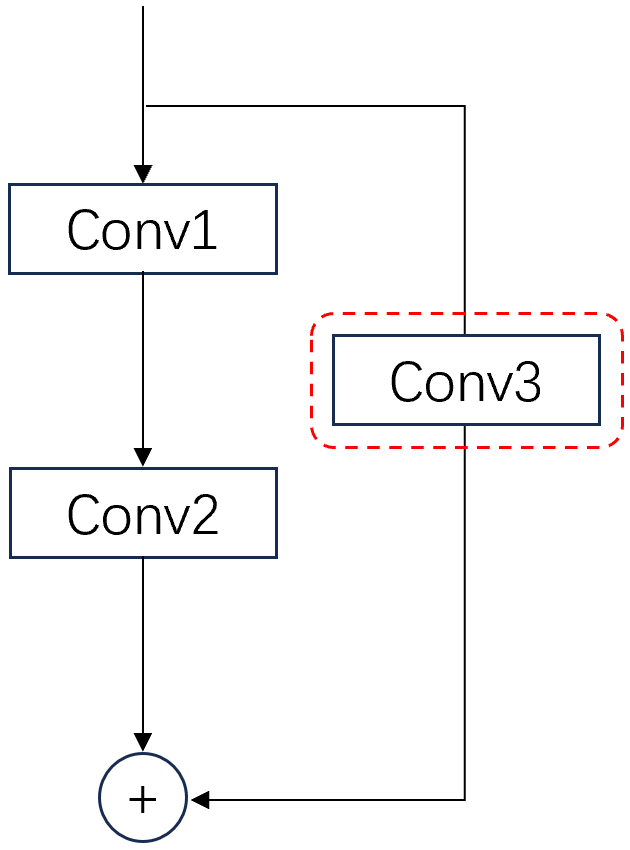}
\caption{Residual structure composed of two or three convolutional layers with skip connections in the residual block. The Res blocks with three convolutional layers have the structure of "Conv3" as shown in the diagram, while the Res blocks with two convolutional layers does not have this structure, the input is directly connected to the output through a skip connection.}
\label{res_block}
\end{figure}

The network commences with a Convolutional (Conv) block, consisting of a single convolutional layer followed by a ReLU activation function. Subsequently, six Residual (Res) blocks follow, each comprising two or three convolutional layers with skip connections, one pooling layer (except the last Res block), and the Leaky ReLU activation function. The architecture of the convolutional layers within the residual block is depicted in Fig.\ref{res_block}.  Finally, a flattening layer is employed to convert the output of the last Residual block into a one-dimensional vector. This flattened vector is then fed into a fully connected neural network for mapping it to a single value.

As elaborated in Section \ref{sec2.3}, in order to make the network convex, some processing should be done on the network. In Fig.\ref{net_stru}, the orange blocks mean that these blocks are not processed, and the blue blocks and green block mean that the weight parameters in these blocks are clipped to retain only the non-negative weight parameters in the convolutional layer. The processed network is convex, detailed principle is illustrated in Section \ref{sec2.3}.

\begin{figure}[!t]
\centering
\includegraphics[width=3.5in]{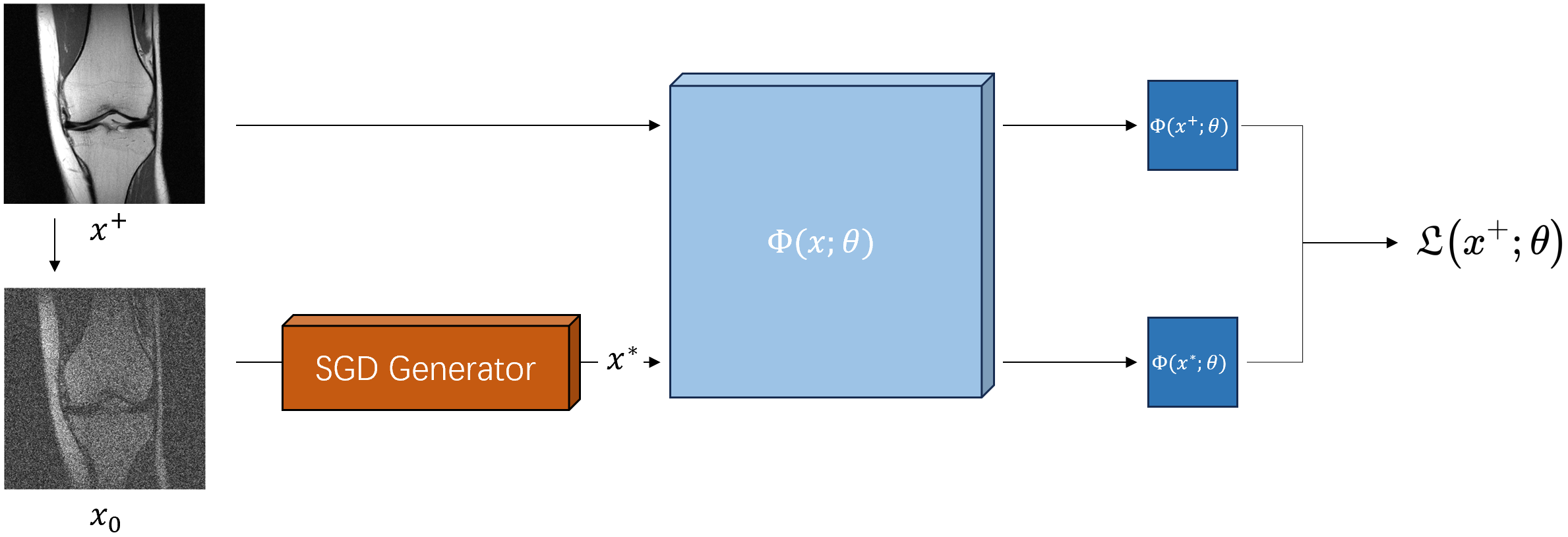}
\caption{Illustration of the training process on network $\Phi \left ( x;\theta \right )$. Real (full-sampled) images $x^{+}$ are corrupted with noise to generate initial images $x_{0}$, and $x_{0}$ are then passed through SGD to generate generated images $x^{*}$. $x^{+}$ and $x^{*}$ are mapped to scalar values through the network $\Phi \left ( x;\theta \right )$ to calculate the loss function.}
\label{whole_process}
\end{figure}

The entire training process of the regularizers is as follows: Real images $x^{+}$ undergo interference through the addition of noise to generate $x_{0}$, followed by the application of stochastic gradient descent method (SGD) to generate the corresponding synthesized samples $x^{*}$. These generated samples, along with the authentic images, are fed into the network and mapped to their respective values. Subsequently, the loss function is calculated and employed to facilitate backpropagation. The entire training process is illustrated in Fig.\ref{whole_process}.

Note that in the training process, we paired the groundtruth with its noisy image as training data. There is no specific training process for different sampling patterns, making it an unsupervised training method. In comparison to supervised reconstruction methods, the proposed approach should exhibit superior robustness and stability.

After the regularizer training was completed, we utilized the Algorithm \ref{alg:pgd} (PGD) to reconstruct the undersampled images, initializing with the zero-filling images. 

\section{Experiment Setup}\label{sec5}
\subsection{Data Acquisition}
The evaluation was conducted using knee and brain MR data with various $k$-space trajectories, including random and uniform cases. The specific details of the MR data are as follows:
\subsubsection{FastMRI Knee data}
The knee raw data \footnote{\url{https://fastmri.org/}} was acquired from a 3T Siemens scanner (Siemens Magnetom Skyra, Prisma and Biograph mMR). Data acquisition used a 15 channel knee coil array and conventional Cartesian 2D TSE protocol employed clinically at NYU School of Medicine. The following sequence parameters were used: Echo train length 4, matrix size $368 \times 368$, in-plane resolution $0.5mm\times0.5mm$, slice thickness $3mm$, no gap between slices. Timing varied between systems, with repetition time (TR) ranging between 2200 and 3000 milliseconds, and echo time (TE) between 27 and 34 milliseconds. We randomly selected 31 individuals (960 slices in total) as training data and 3 individuals (96 slices in total) as test data.
\subsubsection{SIAT Brain data}
The raw data is provided by Shenzhen Institute of Advanced Technology, Chinese Academy of Sciences, acquired from a  3.0T  Siemens  Trio  Tim  MRI  scanner  using  the  T2 weighted  turbo  spin  echo  sequence. The number of coils is 12. The following sequence parameters were used: Echo train length 4, TR/TE = 6100/99 ms, flip angle (FA) = 70°, voxel size = $0.9 \times 0.9 \times 0.9 \text{mm}$, matrix size = $256 \times 232$. We randomly selected 1624 slices to train the net and 157 slices$\left ( \text{from 5 volunteers} \right ) $ to test.
\subsection{Comparative Studies}

Firstly, similar to our proposed CLEAR method, WGAN \cite{arjovsky2017wasserstein} incorporates an adversarial discriminator. Therefore, we have selected WGAN as a comparative method, which is a cutting-edge deep learning technique widely recognized in the field of image processing for its outstanding performance across various image-related tasks. Given that our CLEAR method builds upon AR with latent optimization and convexity enhancements, we have chosen AR as the primary baseline method. Furthermore, based on previous experiences, it is often observed that convex constraints can provide theoretical guarantees to models but may impact the quality of image reconstruction. Hence, we intend to conduct an ablation experiment wherein both the training procedure and testing algorithm remain consistent with CLEAR, with the only difference being the absence of convexity constraints in the network. We refer to this variant as UNCLEAR. Lastly, we will employ the traditional ROF regularization model based on total variation (TV) \cite{rudin1992nonlinear} as a comparative method to evaluate whether our algorithm can match conventional approaches in terms of stability.

\subsection{Training platform}
The deep network models used in this study were implemented on an Ubuntu 20.04 operating system, running on hardware equipped with an NVIDIA A100 Tensor Core GPU with 80 GB of memory. The implementation was carried out using the open-source PyTorch 1.13 framework, which is compatible with CUDA 11.6 and CUDNN for efficient GPU acceleration. This setup allowed for fast and efficient training of the neural network models.

\subsection{Performance Evaluation}
In this study, the quantitative evaluations were all calculated on the image domain. To avoid unfair quantification of the merged images due to inconsistent accuracy of coil sensitivity estimation by different comparison methods, we used ESPIRiT-estimated coil sensitivity merging\cite{uecker2014espirit} for both reference and multi-channel images reconstructed by various methods. For quantitative evaluation, the peak signal-to-noise ratio (PSNR), normalized mean square error (NMSE) value, and structural similarity (SSIM)\cite{wang2004image} index  were adopted.

\section{Results}\label{sec6}

\subsection{Performance Comparison}
\begin{figure}[!t]
\centering
\includegraphics[width=2.5in]{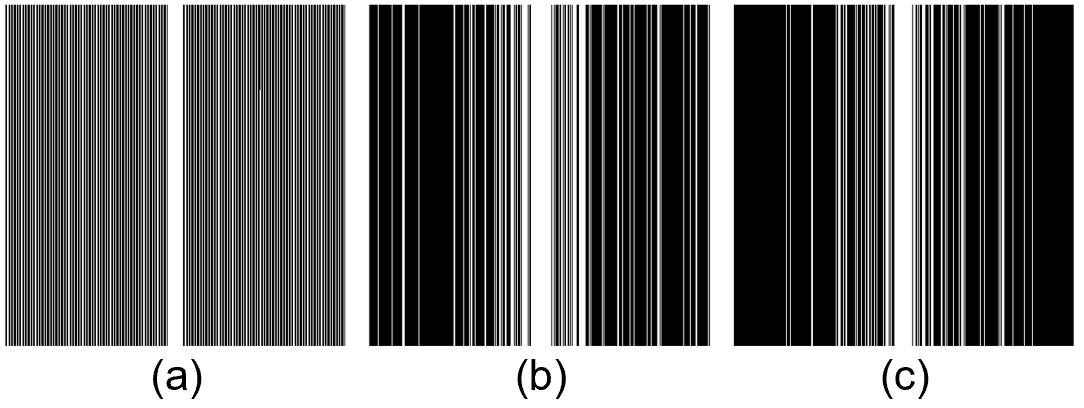}
\caption{1D sampling masks for knee dataset: (a) Uniform sampling mask with an acceleration factor of 3, (b) Random sampling mask with an acceleration factor of 4, (c) Random sampling mask with an acceleration factor of 5.}
\label{knee_mask}
\end{figure}

\begin{figure*}[!t]
\centering
\includegraphics[width=7.0in]{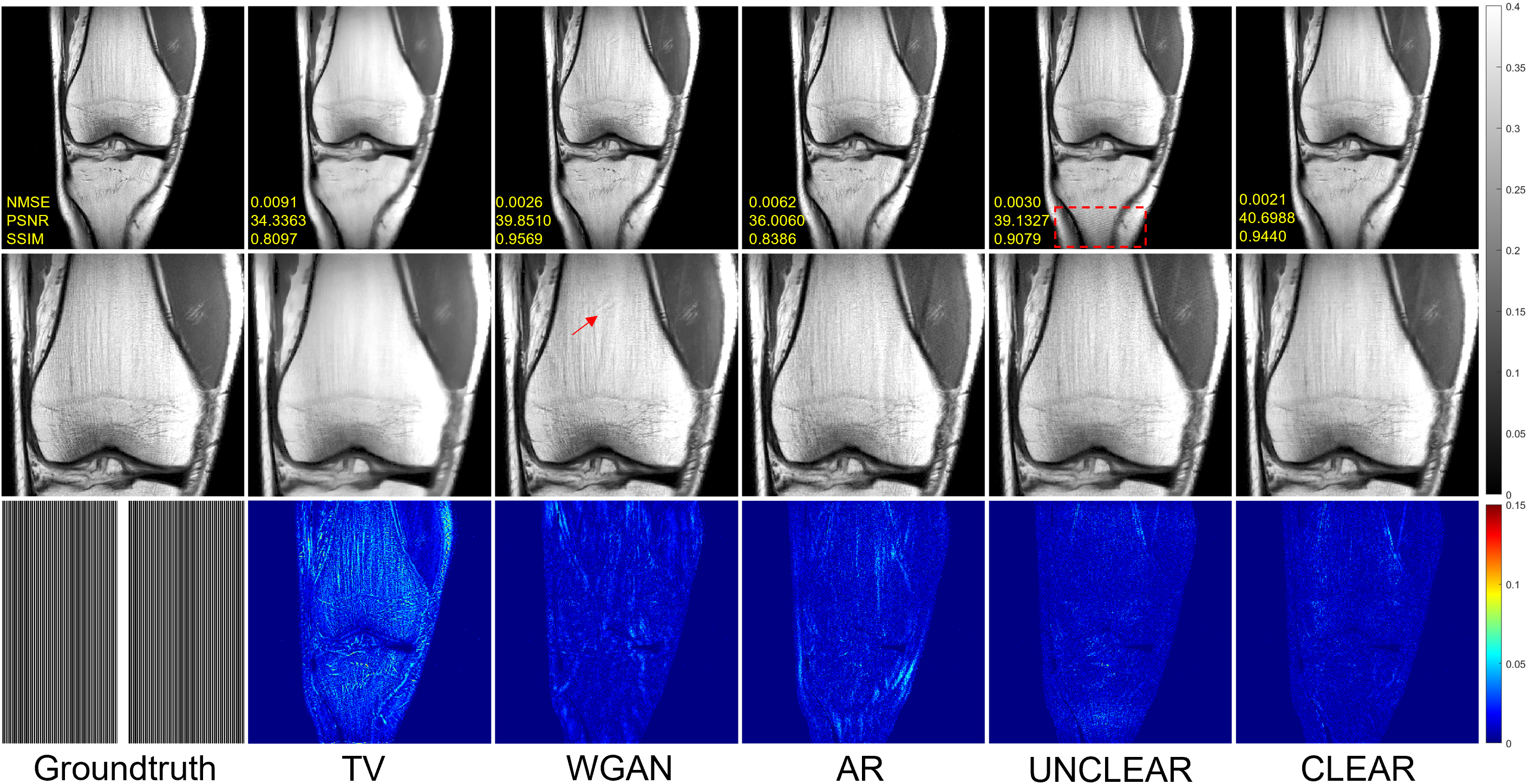}

\caption{Reconstruction results of various comparative methods on the 3-fold undersampled knee data, with the sampling mask shown in the bottom left corner of the figure. CLEAR exhibited the best reconstruction performance.}
\label{uniform3}
\end{figure*}

We initiated our reconstruction experiments under noise-free conditions. For the knee dataset, we employed 1D masks for sampling, encompassing both uniform and random patterns, as illustrated in Fig. \ref{knee_mask}. The acceleration factor for the uniform mask was set to 3. The resulting reconstruction outcomes are presented in Fig. \ref{uniform3}. Notably, the TV method introduced blurriness in fine details, including bone textures. Conversely, the WGAN method exhibited better detail restoration but introduced certain non-existent texture patterns, although they remained relatively inconspicuous at this acceleration factor. The AR reconstruction displayed noticeable aliasing artifacts. In contrast, both UNLEAR and CLEAR demonstrated superior reconstruction performance. However, at careful observation, UNLEAR exhibiting some fish-scale-like distortions at the bottom, which we attribute to the non-convex nature of UNCLEAR's optimization process, leading to some instability. 

\begin{figure*}[!t]
\centering
\includegraphics[width=7.0in]{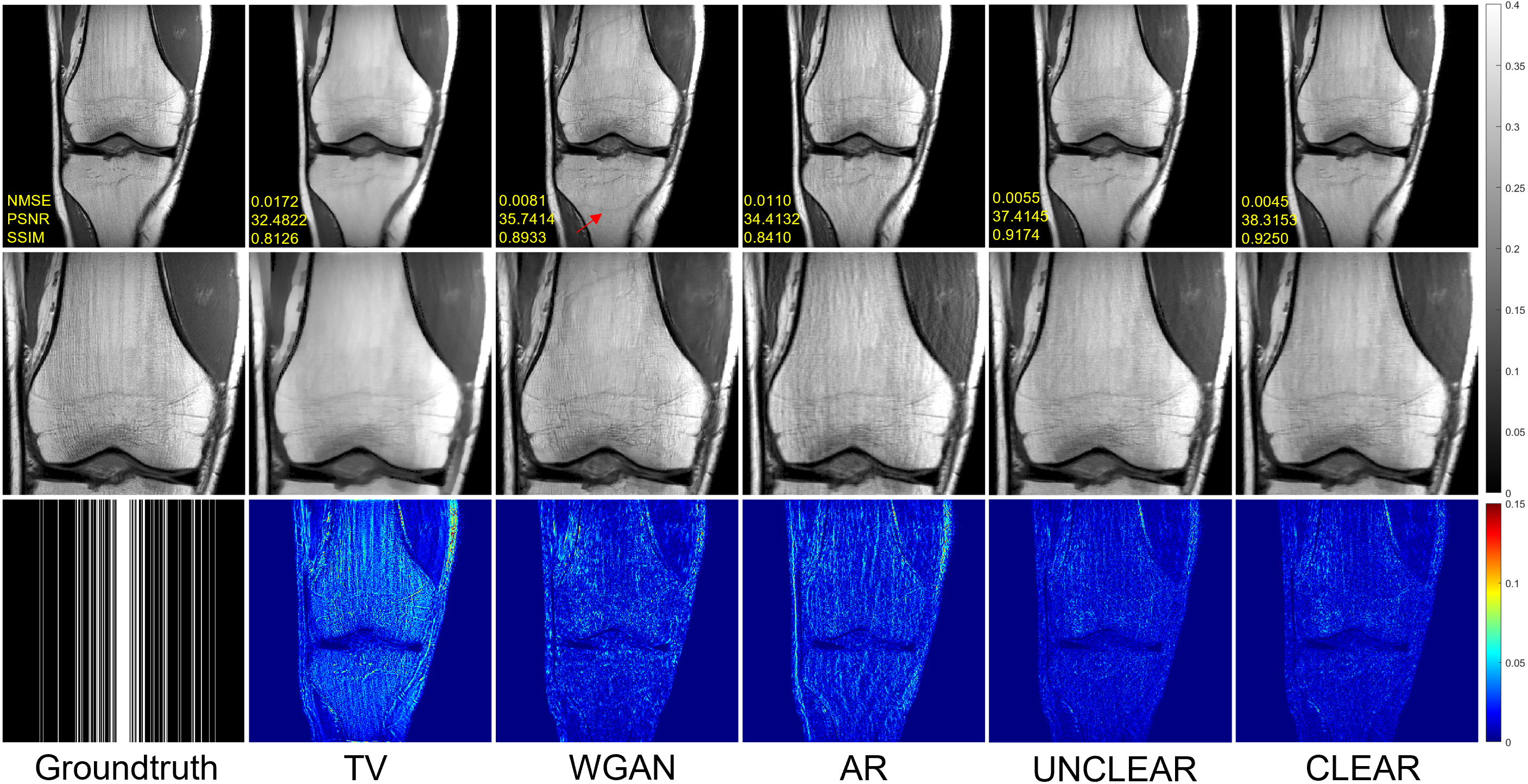}

\caption{Reconstruction results of various comparative methods on the 5-fold undersampled knee data, with the sampling mask shown in the bottom left corner of the figure. CLEAR's recovery result was relatively better.}
\label{cartesian5}
\end{figure*}

The random masks were also employed with acceleration factors of 4 and 5, and the reconstruction outcomes at a 5x acceleration rate are presented in Fig. \ref{cartesian5}. It is evident that at this higher acceleration factor, the TV reconstruction exhibited excessive smoothness due to the limited available information. Conversely, the WGAN reconstruction introduced noticeable, non-existent details and patterns. Both AR and UNCLEAR reconstructions displayed prominent artifacts, with AR exhibiting slightly more severe artifacts. CLEAR's reconstruction performed relatively well, preserving details effectively. We speculate that the reason for the significant degradation of UCLEAR and AR at high acceleration rates lies in the fact that, as acceleration factors increase, the gap between the input zero-filled image and the true image $x^{+}$ widens. This heightened disparity increases the likelihood of non-convex networks becoming trapped in local minima during the iterative process.

Tables \ref{kneeuniform3} to \ref{kneecartesian5} showcase the test metric outcomes for all compared methods on the knee dataset. It is noteworthy that the proposed method consistently outperformed other approaches across various sampling patterns and metrics.
\begin{table}[htbp]
\caption{Test results of various testing methods on the uniformly undersampled knee test dataset, with an acceleration factor of 2.75, highlighting the best results in bold.}
\begin{center}
\begin{tabular}{|c|c|c|c|}
\hline
\textbf{Testing}&\multicolumn{3}{|c|}{\textbf{Evaluation Metrics(Mean±Std)}} \\
\cline{2-4} 
\textbf{Methods} & \textbf{NMSE}& \textbf{PSNR}& \textbf{SSIM} \\
\hline
TV&0.0096±0.0056&32.81±1.93
 &0.7212±0.1307\\
WGAN &0.0065±0.0030&34.37±2.35
 &0.9107±0.0283\\
AR&0.0055±0.0030&35.19±1.85&0.8569±0.0402\\
UNCLEAR&0.0042±0.0026&36.42±2.16&0.8796±0.0441\\
CLEAR&\textbf{0.0035±0.0035}&\textbf{37.52±2.41}&\textbf{0.9145±0.0463}\\
\hline
\end{tabular}
\label{kneeuniform3}
\end{center}
\end{table}

\begin{table}[htbp]
\caption{Test results of various testing methods on the randomly undersampled knee test dataset, with an acceleration factor of 4, highlighting the best results in bold.}
\begin{center}
\begin{tabular}{|c|c|c|c|}
\hline
\textbf{Testing}&\multicolumn{3}{|c|}{\textbf{Evaluation Metrics(Mean±Std)}} \\
\cline{2-4} 
\textbf{Methods} & \textbf{NMSE}& \textbf{PSNR}& \textbf{SSIM} \\
\hline
TV&0.0131±0.0089&31.58±2.18
 &0.6968±0.1269\\
WGAN &0.0074±0.0021&33.53±1.51
 &0.8760±0.0326\\
AR&0.0077±0.0052&33.75±2.00&0.8569±0.0428\\
UNCLEAR&0.0059±0.0046&34.98±2.34&\textbf{0.8893±0.0577}\\
CLEAR&\textbf{0.0053±0.0031}&\textbf{35.27±1.99}&0.8843±0.0527\\
\hline
\end{tabular}
\label{kneecartesian4}
\end{center}
\end{table}

\begin{table}[htbp]
\caption{Test results of various testing methods on the randomly undersampled knee test dataset, with an acceleration factor of 5, highlighting the best results in bold.}
\begin{center}
\begin{tabular}{|c|c|c|c|}
\hline
\textbf{Testing}&\multicolumn{3}{|c|}{\textbf{Evaluation Metrics(Mean±Std)}} \\
\cline{2-4} 
\textbf{Methods} & \textbf{NMSE}& \textbf{PSNR}& \textbf{SSIM} \\
\hline
TV&0.0153±0.0091&30.85±2.11
 &0.6854±0.1270\\
WGAN &0.0100±0.0029&32.23±1.23
 &0.8616±0.0328\\
AR&0.0087±0.0033&32.90±1.36&0.8312±0.0400\\
UNCLEAR&0.0084±0.0062&33.46±2.32&0.8609±0.0529\\
CLEAR&\textbf{0.0071±0.0040}&\textbf{34.05±2.01}&\textbf{0.8720±0.0527}\\
\hline
\end{tabular}
\label{kneecartesian5}
\end{center}
\end{table}

\begin{figure}[!t]
\centering
\includegraphics[width=2.5in]{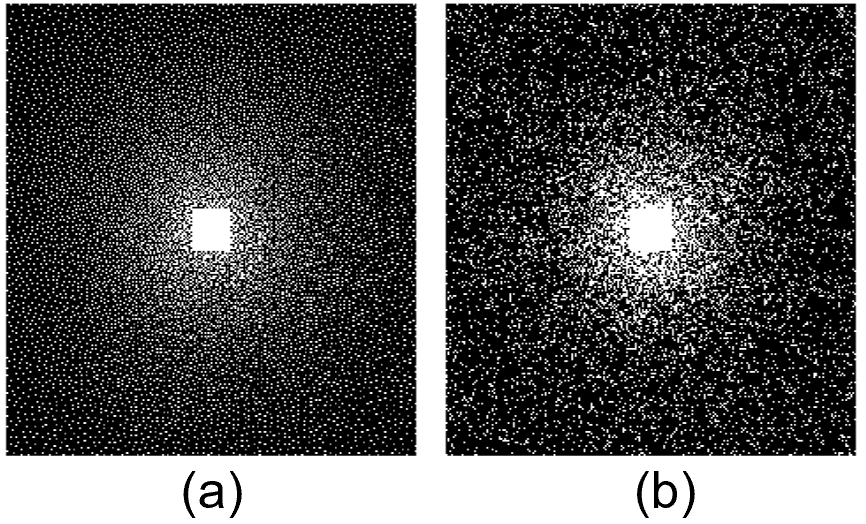}
\caption{2D sampling masks for brain dataset: (a) Poisson random sampling mask with an acceleration factor of 6, (b) Gaussian random sampling mask with an acceleration factor of 6.}
\label{brain_mask}
\end{figure}

\begin{figure*}[!t]
\centering
\includegraphics[width=7.0in]{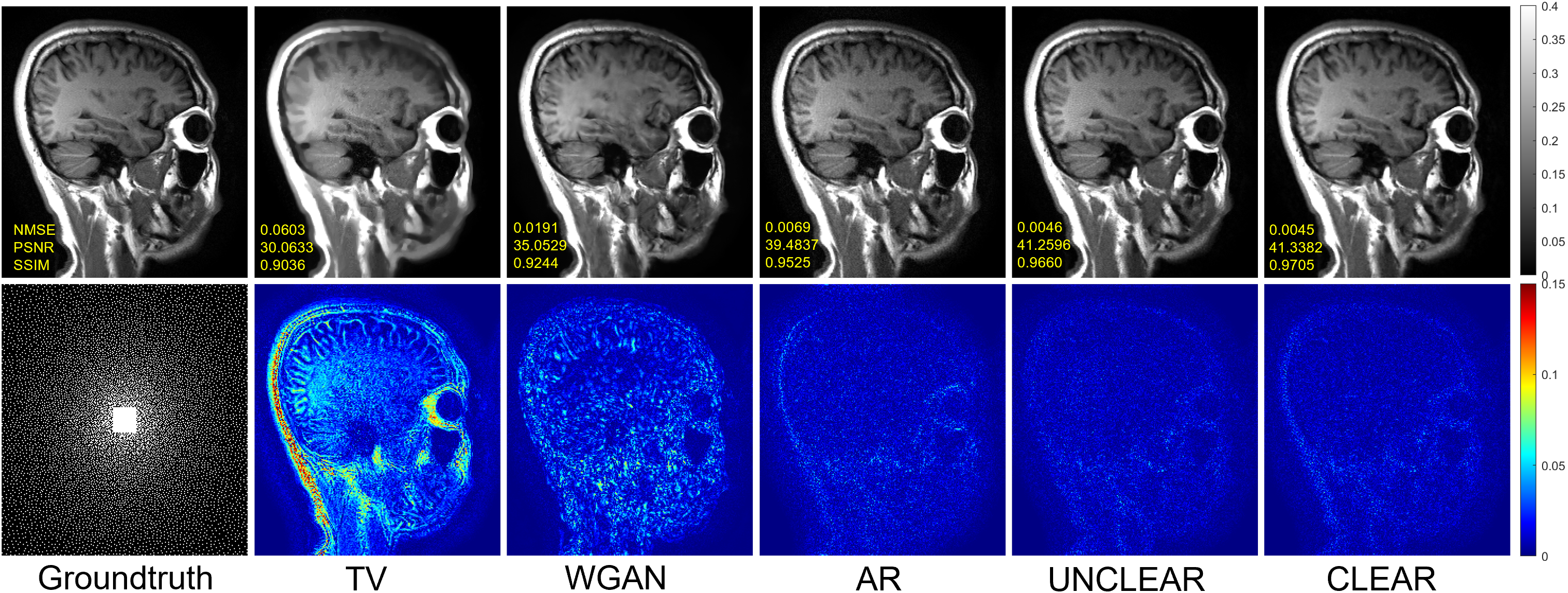}

\caption{Reconstruction results of various comparative methods on the 6-fold undersampled brain data, with the sampling mask shown in the bottom left corner of the figure. CLEAR exhibited the best reconstruction performance.}
\label{possion6}
\end{figure*}

\begin{figure*}[!t]
\centering
\includegraphics[width=7.0in]{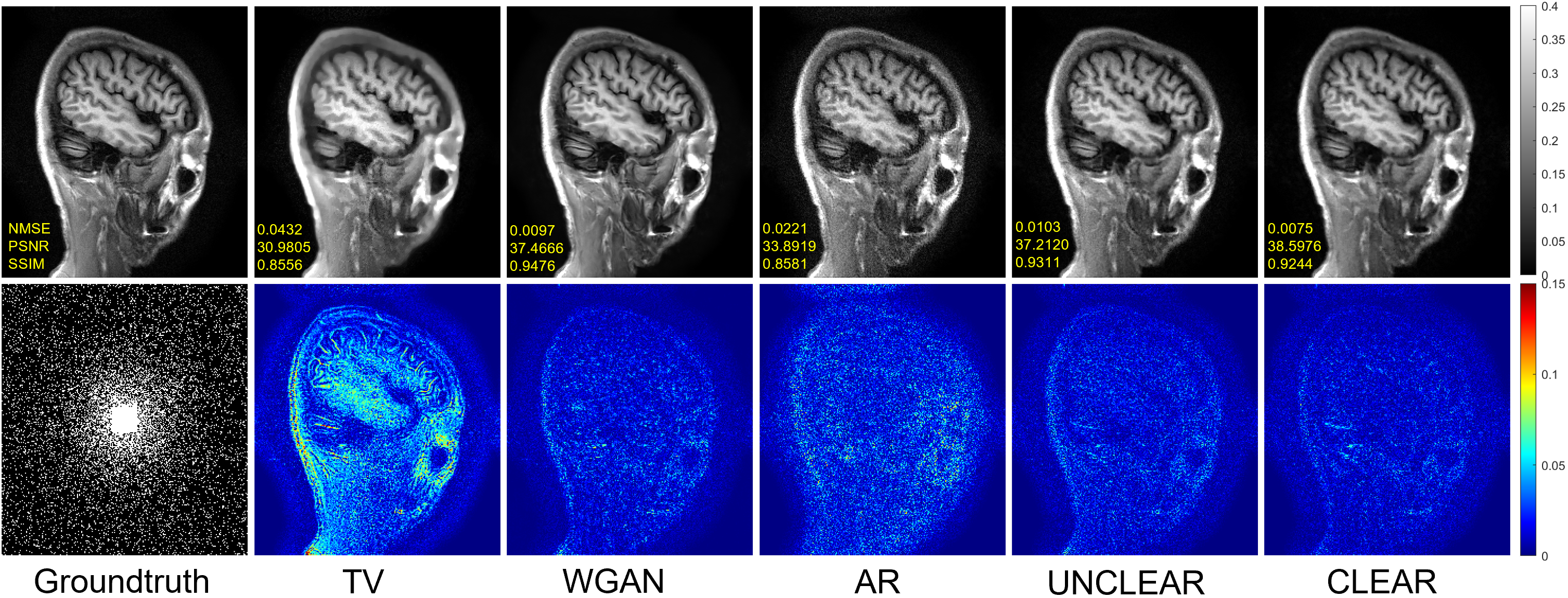}

\caption{Reconstruction results of various comparative methods on the 6-fold undersampled brain data, with the sampling mask shown in the bottom left corner of the figure. CLEAR exhibited the best reconstruction performance.}
\label{random6}
\end{figure*}

In the case of the brain dataset, we utilized 2D masks for the reconstruction process with an acceleration factor of 6, as illustrated in Fig. \ref{brain_mask}. The resulting reconstruction outcomes are depicted in Fig. \ref{possion6} and Fig. \ref{random6}. The TV reconstruction, unfortunately, led to a significant loss of fine details, thereby impacting the overall visual representation of the MR images.
In the case of the WGAN reconstruction, slight distortions within the brain were observed, with this phenomenon being more pronounced in regions featuring intricate contours. This suggests that WGAN may not excel in handling such images, as it tends to introduce non-existent details readily.
The AR-generated images were notably affected by significant noise contamination, compromising their quality. In contrast, both UNCLEAR and CLEAR reconstruction images exhibited commendable performance. However, in terms of visual quality, CLEAR demonstrated a superior signal-to-noise ratio compared to UNCLEAR.

Tables \ref{brainpossion} and \ref{braingau} present the test metric results of all comparison methods on the brain dataset. It was observed that the proposed method achieved the best results in all sampling patterns and metrics.
\begin{table}[htbp]
\caption{Test results of various testing methods on the Possion randomly undersampled brain test dataset, with an acceleration factor of 6, highlighting the best results in bold.}
\begin{center}
\begin{tabular}{|c|c|c|c|}
\hline
\textbf{Testing}&\multicolumn{3}{|c|}{\textbf{Evaluation Metrics(Mean±Std)}} \\
\cline{2-4} 
\textbf{Methods} & \textbf{NMSE}& \textbf{PSNR}& \textbf{SSIM} \\
\hline
TV&0.0281±0.0131&31.24±2.53
 &0.8820±0.0203\\
WGAN &0.0088±0.0043&36.41±2.12
 &0.9372±0.0254\\
AR&0.0059±0.0026&37.91±1.43&0.9365±0.0171\\
UNCLEAR&0.0044±0.0022&39.24±1.60&0.9487±0.0161\\
CLEAR&\textbf{0.0040±0.0016}&\textbf{39.52±1.49}&\textbf{0.9498±0.0163}\\
\hline
\end{tabular}
\label{brainpossion}
\end{center}
\end{table}

\begin{table}[htbp]
\caption{Test results of various testing methods on the Gaussian randomly undersampled brain test dataset, with an acceleration factor of 6, highlighting the best results in bold.}
\begin{center}
\begin{tabular}{|c|c|c|c|}
\hline
\textbf{Testing}&\multicolumn{3}{|c|}{\textbf{Evaluation Metrics(Mean±Std)}} \\
\cline{2-4} 
\textbf{Methods} & \textbf{NMSE}& \textbf{PSNR}& \textbf{SSIM} \\
\hline
TV&0.0342±0.0167&30.47±2.64
 &0.8707±0.0204\\
WGAN &0.0090±0.0040&36.25±1.96
 &0.9370±0.0249\\
AR&0.0080±0.0045&36.69±1.73&0.9271±0.0207\\
UNCLEAR&0.0050±0.0024&38.64±1.54&0.9480±0.0153\\
CLEAR&\textbf{0.0046±0.0019}&\textbf{38.91±1.43}&\textbf{0.9484±0.0158}\\
\hline
\end{tabular}
\label{braingau}
\end{center}
\end{table}

\subsection{Generalization Test}
In this subsection, we conducted generalization tests for the designed CLEAR method and various comparison methods in the context of accelerated MRI.

\begin{figure*}[!t]
\centering
\includegraphics[width=7.0in]{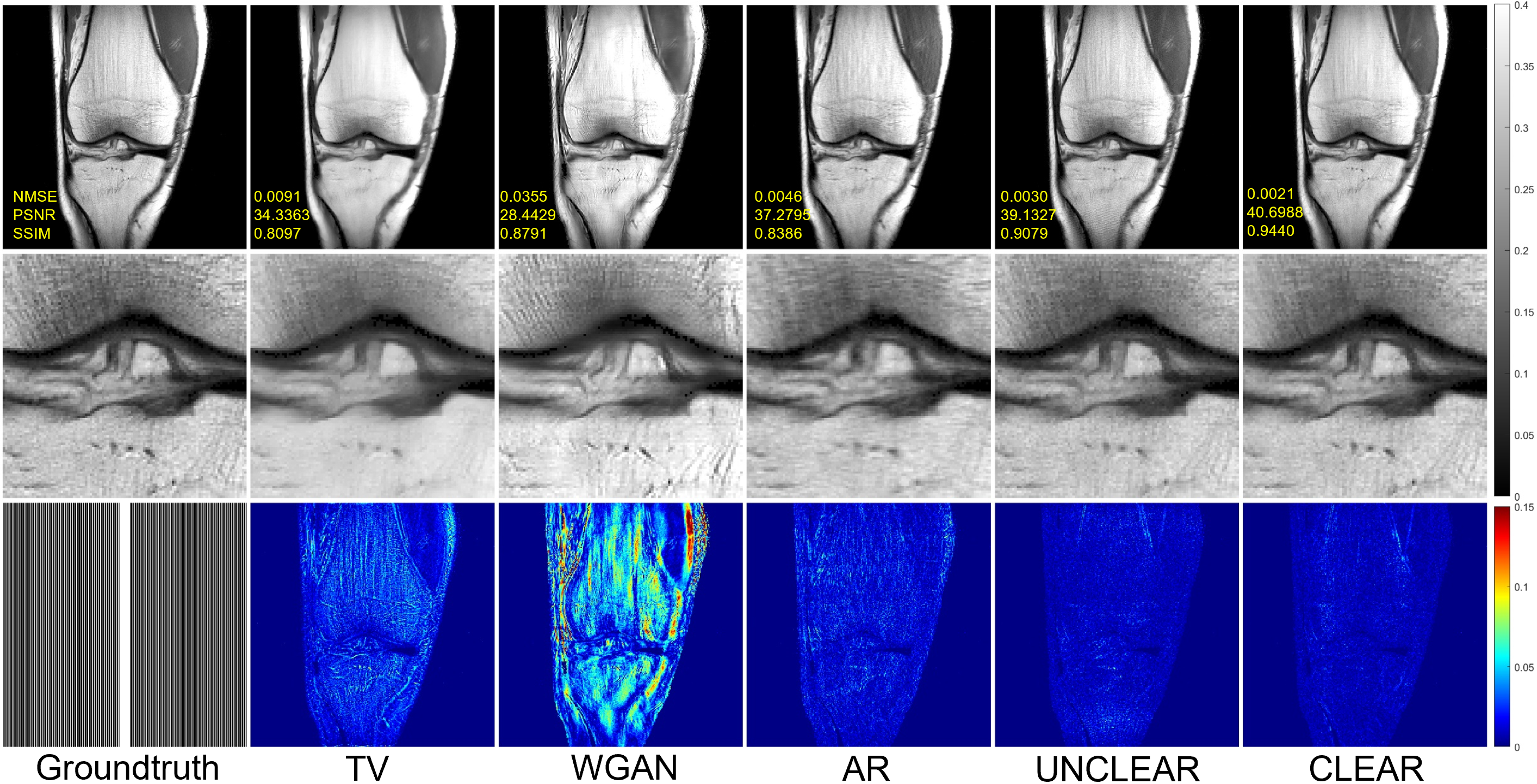}

\caption{The reconstruction results of various comparative methods on the 3-fold undersampled knee data in the mask generalization experiment, with the testing sampling masks displayed in the bottom left corner of the image. Purely data-driven deep learning methods, such as WGAN, exhibited poor performance in terms of generalization.}
\label{mask_gen}
\end{figure*}

Firstly, we performed a sampling pattern shift test on knee data. It is important to note that both CLEAR and UNCLEAR were trained using unsupervised training and were not specifically tailored for any particular sampling pattern. Consequently, after completing training on one dataset, the same pre-trained models were used for reconstruction irrespective of the employed sampling pattern. The sampling pattern generalization experiment proceeded as follows: TV, UNCLEAR, and CLEAR did not undergo any changes in their models since their reconstruction settings remained consistent for any mask pattern. These algorithms were directly tested for reconstruction on the 3-fold-accelerated test dataset, aligning with the corresponding experiments in the previous subsection. On the other hand, WGAN and AR were trained on the knee dataset with a random fourfold undersampling and subsequently tested on a uniformly 3-fold-accelerated knee dataset. The reconstruction results are displayed in Fig. \ref{mask_gen}. It is evident that, except for the three methods that remained unchanged, both WGAN and AR methods exhibited degraded reconstruction results. WGAN, being a purely data-driven deep learning method, encountered task-specific limitations and overfitting, resulting in completely erroneous internal details in its reconstruction due to arbitrary generation. While AR's reconstruction results showed slight improvements in metrics, it also introduced more artifacts in the images.

The test results for mask generalization experiment are presented in Table\ref{tabmaskgen}, with CLEAR achieving the best performance in terms of metrics.
\begin{table}[htbp]
\caption{Test results of various testing methods on the uniformly undersampled knee test dataset in the mask generalization experiment, with an acceleration factor of 3, highlighting the best results in bold.}
\begin{center}
\begin{tabular}{|c|c|c|c|}
\hline
\textbf{Testing}&\multicolumn{3}{|c|}{\textbf{Evaluation Metrics(Mean±Std)}} \\
\cline{2-4} 
\textbf{Methods} & \textbf{NMSE}& \textbf{PSNR}& \textbf{SSIM} \\
\hline
TV&0.0096±0.0056&32.81±1.93
 &0.7212±0.1307\\
WGAN &0.0285±0.0102&27.78±1.46
 &0.8441±0.0272\\
AR&0.0077±0.0051&33.74±2.00&0.8569±0.0429\\
UNCLEAR&0.0042±0.0026&36.42±2.16&0.8796±0.0441\\
CLEAR&\textbf{0.0035±0.0035}&\textbf{37.52±2.41}&\textbf{0.9145±0.0463}\\
\hline
\end{tabular}
\label{tabmaskgen}
\end{center}
\end{table}

\begin{figure*}[!t]
\centering
\includegraphics[width=7.0in]{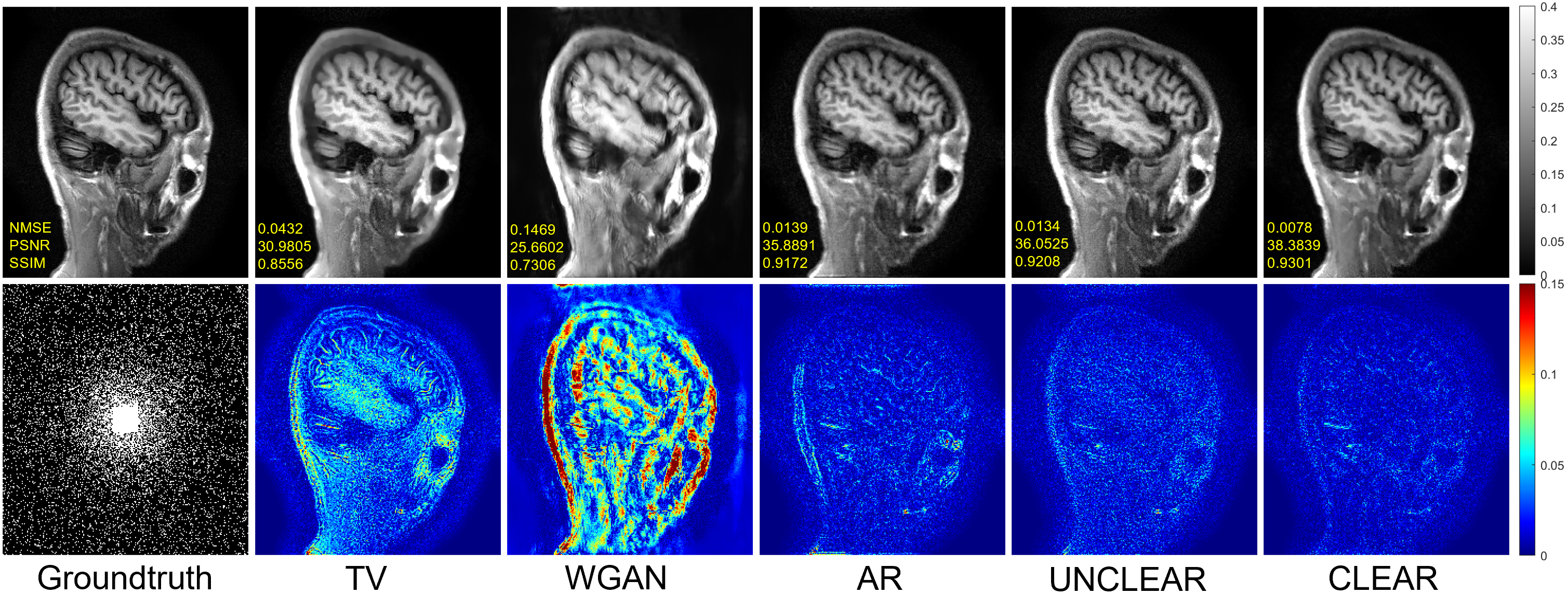}

\caption{The reconstruction results of various comparative methods on the 6-fold undersampled brain data in the dataset generalization experiment, with the testing sampling masks displayed in the bottom left corner of the image. CLEAR exhibited the best generalization performance.}
\label{dataset_gen}
\end{figure*}

Subsequently, we conducted a dataset shift test using a 6-fold-accelerated brain dataset employing Gaussian random sampling masks. With the exception of TV, which does not require specific training, the experimental settings for all other pre-trained models were as follows: WGAN and AR were trained on a 5-fold-accelerated 1D random undersampling knee dataset, while UNCLEAR and CLEAR adopted the unsupervised training on knee dataset described in Section \ref{sec4}.
The reconstruction outcomes are displayed in Fig. \ref{dataset_gen}. TV's reconstruction results remained consistent. However, WGAN, being a purely data-driven deep learning method, exhibited severe distortions in its reconstruction results due to its task-specific limitations, yielding the poorest performance. AR's performance paralleled that of the mask generalization experiment, with some improvement in signal-to-noise ratio metrics and reduced noise contamination but a diminished image quality compared to the results in Fig. \ref{dataset_gen}, leading to increased blurriness. Regarding UNCLEAR, the error map indicated heightened noise contamination and a significant decrease in metrics. Conversely, CLEAR's reconstruction results exhibited minimal changes in both visual and metric aspects compared to Fig. \ref{random6}, showcasing excellent generalization capabilities.

The test results for dataset generalization experiment are presented in Table \ref{tabdatasetgen}, with CLEAR achieving the best performance in all metrics.
\begin{table}[htbp]
\caption{Test results of various testing methods on the Possion randomly undersampled brain test dataset in the dataset generalization experiment, with an acceleration factor of 6, highlighting the best results in bold.}
\begin{center}
\begin{tabular}{|c|c|c|c|}
\hline
\textbf{Testing}&\multicolumn{3}{|c|}{\textbf{Evaluation Metrics(Mean±Std)}} \\
\cline{2-4} 
\textbf{Methods} & \textbf{NMSE}& \textbf{PSNR}& \textbf{SSIM} \\
\hline
TV&0.0342±0.0167&30.47±2.64
 &0.8707±0.0204\\
WGAN &0.0548±0.0280&28.06±1.86
 &0.7100±0.0699\\
AR&0.0082±0.0039&36.31±1.86&0.9209±0.0228\\
UNCLEAR&0.0066±0.0055&37.38±1.79&0.9451±0.0186\\
CLEAR&\textbf{0.0049±0.0018}&\textbf{38.37±1.42}&\textbf{0.9540±0.0146}\\
\hline
\end{tabular}
\label{tabdatasetgen}
\end{center}
\end{table}

\subsection{Robustness Test}
\begin{figure*}[!t]
\centering
\includegraphics[width=7.0in]{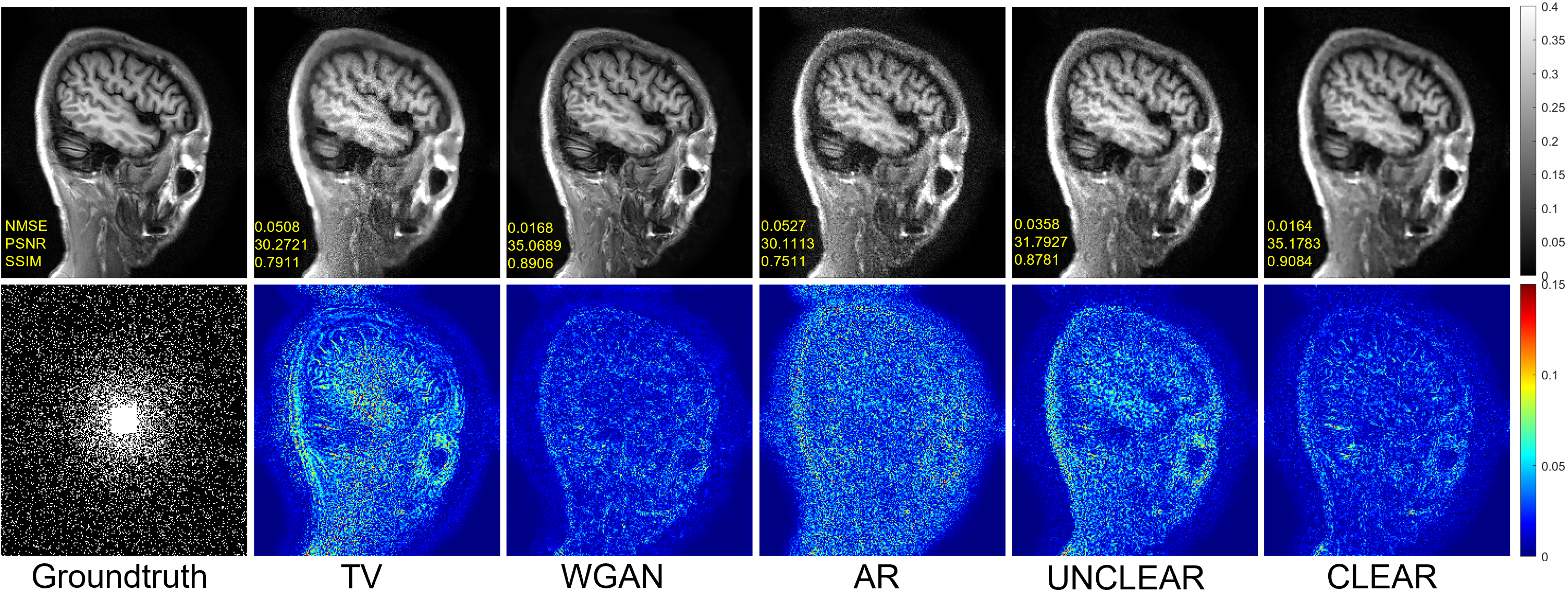}

\caption{The reconstruction results of various comparative methods on the 6-fold undersampled brain data with 50\% Gaussian noise added, with the sampling mask displayed in the bottom left corner. CLEAR achieved the best metric values and exhibited the most stable visual performance, demonstrating superior robustness in handling noisy data.}
\label{robust}
\end{figure*}

Considering the inherent noise contamination during the MRI signal acquisition process, we conducted a robustness test on the reconstruction methods. The test dataset utilized a 6-fold-accelerated brain dataset with Gaussian random undersampling, and 50\% standard Gaussian noise was introduced to its normalized $k$-space data. The training settings were as follows: UNCLEAR and CLEAR were trained using the unsupervised method outlined in Section \ref{sec4}. WGAN and AR were trained using brain data without noise contamination, utilizing the same sampling mask for training as that used for testing.
The reconstruction results are depicted in Fig. \ref{robust}. At this noise level, TV struggled to effectively suppress the noise and lost nearly all details. Although WGAN achieved high metrics, its reconstruction introduced arbitrary textures and details based on noise, resulting in significant deviations from the original images. AR and UNCLEAR exhibited limited robustness against noise contamination, suffering from severe noise interference and a significant decrease in metrics compared to the results in Fig. \ref{random6}. This degradation occurred because the initial values were substantially influenced by noise interference, causing UNCLEAR's iterative solving process to become ensnared in local minima, thus leading to a decline in performance.
CLEAR's reconstruction result was marginally inferior to that in Fig. \ref{random6}, but it still maintained a certain level of quality. It exhibited the best visual effect and metrics among all the compared methods, demonstrating its robustness in handling noise contamination.

Table \ref{tabrobust} presented the test results on the 6-fold undersampled brain data with 50\% Gaussian noise added. The results indicated that CLEAR exhibited the strongest robustness.
\begin{table}[htbp]
\caption{Test results of various testing methods on the Possion randomly undersampled brain test dataset with 50\% Gaussian noise added, with an acceleration factor of 6, highlighting the best results in bold.}
\begin{center}
\begin{tabular}{|c|c|c|c|}
\hline
\textbf{Testing}&\multicolumn{3}{|c|}{\textbf{Evaluation Metrics(Mean±Std)}} \\
\cline{2-4} 
\textbf{Methods} & \textbf{NMSE}& \textbf{PSNR}& \textbf{SSIM} \\
\hline
TV&0.0423±0.0164&29.07±2.00
 &0.7832±0.0335\\
WGAN &\textbf{0.0181±0.0050}&32.60±1.39
 &0.8555±0.0227\\
AR&0.0279±0.0102&30.76±1.37&0.7911±0.0363\\
UNCLEAR&0.0213±0.0058&31.88±1.53&0.8259±0.0408\\
CLEAR&0.0184±0.0064&\textbf{32.61±1.89}&\textbf{0.8774±0.0262}\\
\hline
\end{tabular}
\label{tabrobust}
\end{center}
\end{table}

\begin{figure}[!t]
\centering
\includegraphics[width=3in]{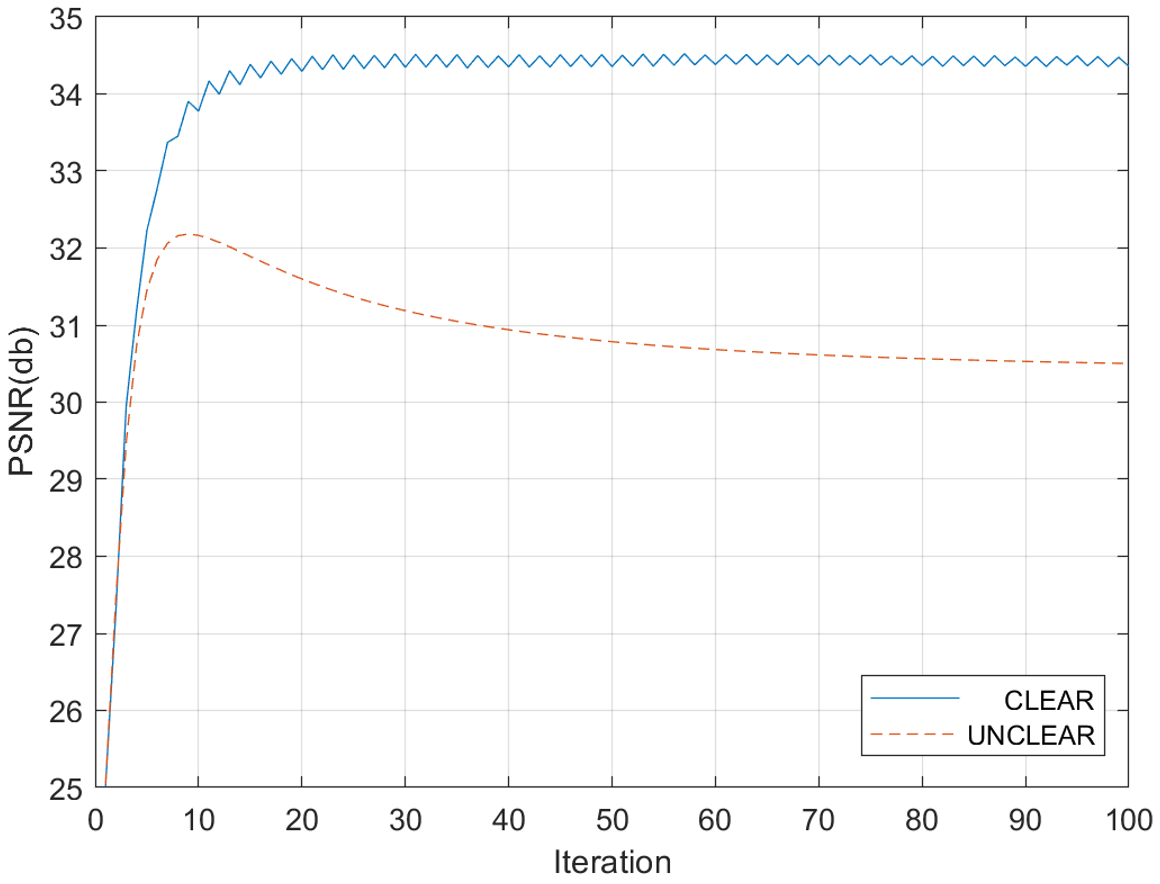}
\caption{The variation of PSNR during 100 iterations of the non-stop reconstruction process using the CLEAR and UNCLEAR methods. CLEAR avoided getting trapped in local minima and thus had a greater advantage in terms of convergence stability. }
\label{converge_plot}
\end{figure}

Figure \ref{converge_plot} illustrates the evolution of PSNR during 100 iterations of the continuous reconstruction process using the CLEAR and UNCLEAR methods when dealing with noisy data in the robustness test mentioned in the preceding paragraph. This graph vividly showcases the convergence advantage of convex networks. The PSNR of the CLEAR method attained optimal convergence with minor fluctuations near the optimum value, whereas the PSNR of the UNCLEAR method failed to converge to the optimal value and displayed substantial fluctuations before ultimately settling into another local minimum. This observation suggests that UNCLEAR, when compared to CLEAR, exhibits increased instability when dealing with noisy data.

\section{Discussion}\label{sec7}
From the previous section, it is evident that CLEAR has demonstrated outstanding performance in terms of both metric results and visual quality when compared to variational regularization methods and state-of-the-art techniques across different datasets and sampling patterns. This superiority becomes even more apparent in the generalization and robustness tests. However, this research still has some limitations.
To ensure convexity, CLEAR employs a specific network structure in which the majority of convolutional kernels are required to be positive. This limitation undoubtedly constrains the network's ability to extract and represent image features fully. It is also a contributing factor to the loss of details in CLEAR's reconstructed images at high acceleration factors. In the CLEAR model, designing a more optimal convex network is a focus for our future work.
Secondly, our current focus has been solely on reconstructing 2D images. In MRI, such as in the case of 2D-t imaging for dynamic cardiac studies, three-dimensional image reconstruction is common. Therefore, within the framework of CLEAR, addressing how to represent the temporal dynamics of images as they evolve over time is also a significant avenue for our future research efforts.

\section{Conclusion}\label{sect8}

In this study, we introduced a CLEAR model, an interpretable deep learning approach designed to address imaging inverse problems. The CLEAR model was capable of fully representing the real data manifold through a set of minima obtained from the learned convex function. We leveraged this learned convex function as a convex regularizer to formulate a CLEAR-informed variational regularization model, which guided the resolution of imaging inverse problems within the real data manifold. Reconstruction was executed using the PGD method on the CLEAR-informed model, which theoretically guaranteed complete reconstruction under specific assumptions. Furthermore, the reconstruction results generated by CLEAR demonstrated theoretical robustness against noise interference. Experimental findings validated CLEAR's substantial performance, generalization capacity, and robustness, establishing its competitiveness in achieving reconstruction results compared to state-of-the-art methods.

{\appendix
\begin{proof}
First, Theorem 3.3 in \cite{cui2022deep} has proved that the distance function $d_{\mathcal{M}}$ is convex and 1-Lipschitz continuous. Then we will prove that $\Omega = \mathcal{M}$, a.e. On the one hand, we have
\begin{equation*}\begin{aligned}
& \mathbb{E} _{x^{*} \in \Omega }   \left [  f\left (  x^{*}\right )\right]- \mathbb{E} _{ x^{+}\sim \mathbb{P} _{x} }\left[ f\left (  x^{+}\right ) \right ] \\& = \mathbb{E}_{x^{+}\sim \mathbb{P} _{x} }  \left [  f_{x^{*} \in \Omega }\left (  x^{*}\right )-f\left (  x^{+}\right ) \right ]\le 0\\
\end{aligned}\end{equation*}
On the other hand, we have
\begin{equation*}\begin{aligned}
&\mathbb{E} _{x^{*} \in \Omega }  \left [  f\left (  x^{*}\right )\right]- \mathbb{E} _{ x^{+}\sim \mathbb{P} _{x} }\left[ f\left (  x^{+}\right ) \right ] \\& \ge   \mathbb{E}_{x^{*} \in \Omega }  \left [  d_\mathcal{M} \left (  x^{*}\right )\right]- \mathbb{E} _{ x^{+}\sim \mathbb{P} _{x} }\left[ d_\mathcal{M}\left (  x^{+}\right ) \right ]\\
=&\mathbb{E}_{x^{*} \in \Omega }  \left [  d_\mathcal{M} \left (  x^{*}\right )\right]\ge 0
\end{aligned}\end{equation*}
According to  inequalities above, we have,
\begin{equation*}\begin{aligned}
&\mathbb{E}_{x^{+}\sim \mathbb{P} _{x} }  \left [  f_{x^{*} \in \Omega }\left (  x^{*}\right )-f\left (  x^{+}\right ) \right ] = 0,\mathbb{E}_{x^{*} \in \Omega }  \left [  d_\mathcal{M} \left (  x^{*}\right )\right]=0
\end{aligned}\end{equation*}
Since $f\left (  x^{+}\right ) \ge f_{x^{*} \in \Omega }\left (  x^{*}\right )$, we have
\begin{equation*}\begin{aligned}
&f_{x^{+} \in \mathcal{M} }\left (  x^{+}\right ) = \min_{x\in X} f\left (  x\right ) \ a.e.,\quad d_\mathcal{M} \left (  x^{*}\right )=0 \ a.e.
\end{aligned}\end{equation*}
The former equality leads to $\mathcal{M}\subseteq \Omega \ a.e.$, and the later equality leads to $\Omega \subseteq \mathcal{M} \ a.e.$, and we prove that $\Omega  =  \mathcal{M} \ a.e.$
\end{proof}
\begin{proof} 
The subgradient of $f$ is defined as
\begin{equation*}\begin{aligned}
&\partial f\left ( x \right ) =\left \{ g\in \mathbb{R}^{n }|\left \langle g,v \right \rangle \le f\left ( x+v \right ) -f\left ( x \right ) ,\forall  v\in \mathbb{R}^{n } \right \}  
\end{aligned}\end{equation*}

Define $\mathcal{A} = \left \{ x\in X | Ax = b\right \} $, and it's obviously $\mathcal{A}$ is a convex set.

Denoting the projection of $x$ on $\mathcal{A}$ by $x'$,we first prove that for $\forall z \in  X$ and $\forall x \in  \mathcal{A} $ ,the following inequality is always satisfied,
\begin{equation}\begin{aligned}\label{ineq1}
& \left \langle z'-z,z'-x \right \rangle \le 0 
\end{aligned}\end{equation}
because $\mathcal{A} $ is convex and $z'$ is the projection of z, we have $\forall t > 0 $
\begin{equation*}\begin{aligned}
& \left \| z'+t(x-z')-z \right \|^{2} \ge \left \| z'-z \right \|^{2} 
\end{aligned}\end{equation*}
unfolding the equality, we get,
\begin{equation*}\begin{aligned}
& \left \langle z'-z,x-z' \right \rangle+ \frac{t}{2} \left \| x-z' \right \| ^{2} \ge 0
\end{aligned}\end{equation*}
when $t \to 0$, we get the inequality.

Then we prove that $\forall y,z \in X$, the distance of their projection on A is nearer than themself,
\begin{equation*}\begin{aligned}
& \left \| y'-z' \right \|  \le  \left \| y-z \right \| 
\end{aligned}\end{equation*}
according to the inequality (\ref{ineq1}),
\begin{equation*}\begin{aligned}
& \left \langle z-z',y'-z' \right \rangle \le 0 ,\ \left \langle y-y',y'-z' \right \rangle \ge 0
\end{aligned}\end{equation*}
so that,
\begin{equation*}\begin{aligned}
& \left \langle y-y'-\left ( z-z' \right ) ,y'-z' \right \rangle \ge 0\\
&\left \langle y-z-\left ( y'-z' \right ) ,y'-z' \right \rangle \ge 0\\
&\left \langle  y'-z'  ,y'-z' \right \rangle \le \left \langle y-z ,y'-z' \right \rangle \\
&\left \| y'-z' \right \| ^{2}\le \left \langle y-z ,y'-z' \right \rangle \le \left \| y-z \right \| \left \| y'-z' \right \| \\
&\left \| y'-z' \right \|\le \left \| y-z \right \|  
\end{aligned}\end{equation*}
we have proved the conclusion.

Supposing $\mathcal{N}=\left \{ x|x\in \mathcal{A} \cap \mathcal{M} \right \} $, obviously $x^{+}\in  \mathcal{N} $
\begin{equation*}\begin{aligned}
x_{k+0.5} &= x_{k}-t_{k} \ast \partial f\left ( x_{k} \right )\\
x_{k+1} &= \mathcal{P}_{\mathcal{A}}(x_{k+0.5}) \\
\left \|  x_{k+1}-x^{+}\right \|^{2}&\le \left \|x_{k+0.5} -x^{+} \right \| ^{2}\\
&= \left \|x_{k}-t_{k} \ast \partial f\left ( x_{k} \right ) -x^{+} \right \| ^{2}\\
&=\left \| x_{k} -x^{+}\right \| ^{2} + t_{k}^{2} \left \| \partial f\left ( x_{k} \right ) \right \| ^{2}\\&-2t_{k}\left \langle x_{k} -x^{+},\partial f\left ( x_{k} \right ) \right \rangle\\
2t_{k}\left \langle x_{k} -x^{+},\partial f\left ( x_{k} \right ) \right \rangle&\le \left \| x_{k} -x^{+}\right \| ^{2} + t_{k}^{2} \left \| \partial f\left ( x_{k} \right ) \right \| ^{2}\\&-\left \|  x_{k+1}-x^{*}\right \|^{2}
\end{aligned}\end{equation*}
according to the definition of the subgradiant and the 1-Lipschitz of f, we can acquire the inequation,
\begin{equation*}\begin{aligned}
2t_{k}\left ( f\left ( x_{k} \right )  -f\left (  x^{+}\right ) \right ) &\le \left \| x_{k} -x^{+}\right \| ^{2} -\left \|  x_{k+1}-x^{+}\right \|^{2}+t_{k}^{2}, \\&\quad \quad \quad \quad \quad \quad \quad \quad \quad \quad \quad k=0,1,\dots
\end{aligned}\end{equation*}
sum both sides separately, we get,
\begin{equation*}\begin{aligned}
2\sum_{0}^{k} t_{k}\left ( f\left ( x_{k} \right )  -f\left (  x^{+}\right ) \right ) &\le \left \| x_{0} -x^{+}\right \| ^{2} +\sum_{0}^{k}t_{k}^{2}
\end{aligned}\end{equation*}
A proper choice of $t_{k}$ will allow $ f\left ( x_{k} \right )$ to converge to $f\left (  x^{+}\right )$.

We have proved that $f\left ( x_{k} \right ) \to f\left (  x^{+}\right )$, next we will prove that $x_{k} \to x^{+}$.

Denote the distance from $x$ to $\mathcal{N}$ as $d_{\mathcal{N}}\left ( x \right ) $, the following proof will prove that $d_{\mathcal{N}}\left ( x_{k} \right ) \to 0$.

Assuming $d_{\mathcal{N}}\left ( x_{k} \right )$ do not converge to $0$, this means that $\exists \epsilon$,  there is always $d_{\mathcal{N}}\left ( x_{k} \right ) \ge \epsilon$, $k>k_{t}$, no matter how large $k_{t}$ is taken. We can easily get a subsequence of $\left \{  x_{k}\right \} $, denoted $\left \{  x_{k'}\right \}$, for each term in$\left \{  x_{k'}\right \}$ , there's always $d_{\mathcal{N}}\left ( x_{k'}\right ) \ge \epsilon$.

Let the projection $\mathcal{P}_{\mathcal{N}}(x_{k'})$ of $x_{k'}$ onto $\mathcal{N}$ be $x'_{k'}$. We get a new sequence $\left \{ x''_{k'} \right \} $ , where  $x''_{k'} $ lies on the line between $x_{k'}$ and $x'_{k'}$ and $d\left ( x''_{k'},x'_{k'} \right )$ is $\epsilon$, i.e., $\left \| x''_{k'}-x'_{k'}  \right \| = \epsilon$. It's easy to prove that  $\mathcal{P}_{\mathcal{N}}(x''_{k'}) = x'_{k'}$, which imply that $d_{\mathcal{N}}\left ( x''_{k'}\right ) = \epsilon$. Set $\mathcal{M}^{+} = \left \{ x|x \in X, d_{\mathcal{M}}\left ( x\right )\le \epsilon \right \}  $. Considering that $d_{\mathcal{M}}$ is a 1-Lip continuous convex function, and $x \in X \subseteq \mathbb{R} ^{N}$, we get that $\mathcal{M}^{+}$ is a bounded closed set. In the $\mathbb{R} ^{N}$ space, as the suquence of points in $\mathcal{M}^{+}$,  $\left \{ x''_{k'} \right \} $must have a convergent subsequence, denoted  $\left \{ x''_{k''} \right \} $.

Denoted the point $\left \{ x''_{k''} \right \} $ converge to by $x''_{*}$. $\left \{  x_{k'}\right \}$ is the subsequence of $\left \{  x_{k}\right \}$, and $\left \{  x_{k}\right \}$ is generated by projection operator, as a result, $\left \{  x_{k'}\right \} \subseteq \left \{  x_{k}\right \} \subseteq \mathcal{A}$. On the other hand, $\left \{  x'_{k'}\right \} \subseteq \mathcal{N} \subseteq \mathcal{A} $. And, as mentioned above, $\mathcal{A}$ is a convex set. Then we can get that $\left \{  x''_{k'}\right \}  \subseteq \mathcal{A} $. $\left \{  x''_{k''}\right \}$ is the subsequence of $\left \{  x''_{k'}\right \}$, the following conclusion follow naturally that $\left \{  x''_{k''}\right \} \subseteq \mathcal{A}$, i.e., $Ax''_{k''}=b$ for each term in $\left \{  x''_{k''}\right \}$. It's easy to prove that $Ax''_{*}=b$, i.e., $x''_{*} \in \mathcal{A}$.

In the previous part of the proof, we have proved that $ f\left ( x_{k} \right ) \to f\left (  x^{+}\right )$. So, $ f\left ( x_{k'} \right ) \to f\left (  x^{+}\right )$. And, obviously, $f\left ( x'_{k'} \right ) = f\left (  x^{+}\right )$. Then, $f$ is a convex function, for $x''_{k'}$ lies on the line between $x_{k'}$ and $x'_{k'}$, there is $ f\left ( x''_{k'} \right ) \le f\left (  x_{k'}\right )$. So, $ f\left ( x''_{k'} \right ) \to f\left (  x^{+}\right )$, naturally, $ f\left ( x''_{k''} \right ) \to f\left (  x^{+}\right )$. Because $f$ is a 1-Lip continuous function, $ f\left ( x''_{k''} \right ) \to f\left (  x^{+}\right )$, $x''_{k''} \to x''_{*}$, we can get $ f\left ( x''_{*} \right ) = f\left (  x^{+}\right )$, i.e., $x''_{*} \in \mathcal{M}$. Combined with the last paragraph, we get that $x''_{*} \in \mathcal{N}$.

As the subsequence of $\left \{ x''_{k'} \right \}$, $\left \{ x''_{k''} \right \}$ keep the property that $d_{\mathcal{N}}\left ( x''_{k''}\right ) = \epsilon$. However, combined with the previous derivation, there is $d_{\mathcal{N}}\left ( x''_{k''}\right ) \le d\left ( x''_{k''},x''_{*} \right ) \to 0$. This is a contradiction and shows that the assumption at begin was wrong. So, we prove that $d_{\mathcal{N}}\left ( x_{k} \right ) \to 0$. Under assumption2, $\mathcal{N} = \left \{ x^{+} \right \} $, so, $d\left ( x_{k},x^{+} \right ) = d_{\mathcal{N}}\left ( x_{k} \right ) \to 0 $, i.e., $x_{k} \to x^{+}$.
\end{proof}
\begin{proof}
The proof of Theorem \ref{thm5}. Under Assumptions \ref{assup1} and \ref{assup2}, let's consider the scenario where the clean image is contaminated by noise $\delta$. The measurement result is denoted as $b_{\delta}$, and the corresponding iteration result after a substantial number of steps is denoted as $x^{k}_{\delta}$. Then, $x^{k}_{\delta}$ is continuous at $0$ with respect to $\delta$, meaning that as $\delta$ approaches $0$, $x^{k}_{\delta}$ approaches the true solution $x^{+}$ when $k$ is sufficiently large, i.e., $x_{\delta}\to x^{+}$, when $\delta \to 0, k\to\infty$.

Set $\mathcal{A}_{\delta} = \left \{ x|Ax = b+\delta  \right \} $, then set $\mathcal{N}_{\delta} = \left\{x| x =\arg\min\limits_{x\in \mathcal{A}_{\delta} }f\left(x\right) \right\}$. Denote the projection of $x^{+} = \mathcal{A} \cap \mathcal{M}$ onto $\mathcal{A}_{\delta}$ by $x_{\delta}$. Give following conclusion without proof,
\begin{equation*}\begin{aligned}
x_{\delta} \to x^{+}, \text{when} \delta \to 0
\end{aligned}\end{equation*}
Set $d_{\delta} = \sup\limits_{x \in \mathcal{N}_{\delta}}d\left(x,x_{\delta}\right)$, if $d_{\delta}$ does not have a concrete supremum, let $d_{\delta}=\infty $. We first prove $d_{\delta} \to 0$. If $d_{\delta}$ does not converge to $0$ when $\delta \to 0$, that means, $\exists \epsilon$, there's always $d_{\delta'} \ge \epsilon,\left \| \delta' \right \| \le \left \|  \delta \right \|  $, no matter how little $\left \|  \delta \right \|$ is.

$d_{\delta} \ge \epsilon$ means there exists one point $ x'_{\delta} \in \mathcal{A}_{\delta}$ with following property, $d\left(x'_{\delta},x_{\delta}\right) \ge \epsilon$ and $f\left(x'_{\delta}\right) \le f\left(x_{\delta}\right)$. Like how to get $x''_{k'}$ in last proof, we get a point $x''_{\delta}$ from the line between $x'_{\delta}$ and $x_{\delta}$ and make $d\left(x''_{\delta},x_{\delta}\right)=\epsilon$. Because $f$ is a convex function and $\mathcal{A}_{\delta}$ is a convex set, we get $x''_{\delta} \in \mathcal{A}_{\delta}$ and $f\left(x''_{\delta}\right) \le f\left(x_{\delta}\right)$.

If $d_{\delta}$ does not converge to $0$ when $\delta \to 0$, we can construct a sequence $\left \{ \delta_{k} \right \} $, let$\left \{ \delta_{k} \right \} \to 0$ and $d_{\delta_{k}}\ge \epsilon$. The points sequence$\left \{ x''_{\delta_{k}} \right \} $ is obviously in a bounded closed set, so it must have a  convergent subsequence $\left \{ x''_{\delta_{k'}} \right \} $. Denote the point $\left \{ x''_{\delta_{k'}} \right \} $ converge to by $x^{*}$. Corresponding projection sequence of $x^{+}$ onto $\mathcal{A}_{\delta_{k'}}$ is $\left \{ x_{\delta_{k'}} \right \} $, here we should note that $d\left(x''_{\delta_{k'}},x_{\delta_{k'}}\right)=\epsilon$ and $\left \{ x_{\delta_{k'}} \right \} \to x^{+}$. Because $Ax''_{\delta_{k'}}=b+\delta_{k'} \to b$, $x''_{\delta_{k'}}\to x^{*}$, there is $Ax^{*}=b$, i.e., $x^{*}\in \mathcal{A}$. On the other hand, $f\left(x^{+}\right) \le f\left(x''_{\delta_{k'}}\right) \le f\left(x_{\delta_{k'}}\right) \to f\left(x^{+}\right)$, we can get $f\left(x''_{\delta_{k'}}\right) \to f\left(x^{+}\right)$. And $x''_{\delta_{k'}}\to x^{*}$, there is $f\left(x^{*}\right) = f\left(x^{+}\right)$, i.e., $x^{*}\in \mathcal{M}$. As a result, $x^{*}\in \mathcal{M} \cap \mathcal{A}$. According to assumption 2, $x^{+} \in \mathcal{M} \cap \mathcal{A}$ is unique, that means $x^{*}=x^{+}$. Then we have $d\left(x''_{\delta_{k'}},x_{\delta_{k'}}\right) \le d\left(x''_{\delta_{k'}},x^{*}\right) + d\left(x^{+},x_{\delta_{k'}}\right) \to 0$. This is a contradiction and shows that the assumption at begin was wrong, i.e., $d_{\delta}$ surely converges to $0$ when $\delta \to 0$.

Denote the iteration solution of projection subgradient descent method to following problem by $x^{k}_{\delta}$, 
\begin{equation*}\begin{aligned}
\arg\min\limits_{x\in \mathcal{A}_{\delta} }f\left(x\right).
\end{aligned}\end{equation*}
In the proof of theorem 3.4 (i.e. last proof), we have proved that $d_{\mathcal{N}_{\delta}}\left(x^{k}_{\delta}\right) = d\left(x^{k}_{\delta}, \mathcal{P}_{\mathcal{N}_{\delta}}\left(x^{k}_{\delta}\right)\right) \to 0$,when $k \to \infty$. We finally get following inequality,
\begin{equation*}\begin{aligned}
d\left(x^k_{\delta},x^{+}\right) \le d_{\mathcal{N}_{\delta}}\left(x^{k}_{\delta}\right) + d_{\delta} + d\left(x_{\delta},x^{+}\right) &\to 0, \\ \text{when} \delta \to 0, k &\to \infty
\end{aligned}\end{equation*}
\end{proof}
}


\bibliographystyle{IEEEtran}
\bibliography{IEEEabrv,ref}

\vfill

\end{document}